\pdfoutput=1

\documentclass[11pt]{article}

\usepackage{acl}

\usepackage{times}
\usepackage{latexsym}

\usepackage[T1]{fontenc}

\usepackage[utf8]{inputenc}

\usepackage{microtype}

\usepackage{inconsolata}

\usepackage{soul}
\usepackage{url}
\usepackage{graphicx}
\usepackage{tabularx}
\usepackage{amsmath}
\usepackage{amsthm}
\usepackage{booktabs}
\usepackage{algorithm}
\usepackage{algorithmic}
\usepackage{subcaption}
\usepackage{xcolor}
\usepackage{comment}
\usepackage{enumitem}
\usepackage{multicol}
\usepackage{multirow}
\usepackage{CJKutf8}

\newcommand{\zhsmall}[1]{\begin{CJK*}{UTF8}{gbsn}{#1}\end{CJK*}}

\newcommand{\datasetidioms}{IDIOM}
\newcommand{\datasettp}{TangPoetry}
\newcommand{\datasetpn}{ProperNoun}
\newcommand{\datasetterm}{Terminology}
\newcommand{\datasetpopqa}{PopQA}
\newcommand{\datasetlama}{LAMA-UHN}
\newcommand{\datasetcele}{CelebrityParent}

\newcommand{\llama}{LLaMA}
\newcommand{\qwen}{Qwen}
\newcommand{\mistral}{Mistral}
\newcommand{\gemma}{Gemma}
\newcommand{\gptfour}{GPT-4}

\newcommand{\modelname}{{ROME} }

\title{\modelname: Memorization Insights from Text, Logits and Representation}
\author{Bo Li $^{1,4}$\thanks{Equal contribution.}, Qinghua Zhao$^{2,3}$ \footnotemark[1], Lijie Wen$^1$ \thanks{The corresponding author.}\\
$^1$Tsinghua University, Beijing, China  
$^2$Beihang University, Beijing, China \\
$^3$ University of Copenhagen, Copenhagen, Denmark 
$^4$ Baidu Inc., Beijing, China \\
\texttt{li-b19@mails.tsinghua.edu.cn, zhaoqh@buaa.edu.cn, wenlj@tsinghua.edu.cn
}
}

\begin{document}
\maketitle

\begin{abstract}
Previous works have evaluated memorization by comparing model outputs with training corpora, examining how factors such as data duplication, model size, and prompt length influence memorization.  However, analyzing these extensive training corpora is highly time-consuming. To address this challenge, this paper proposes an innovative approach named \modelname~\footnote{\textbf{\modelname~} refers to the four letters in ``m\textbf{emor}ization'', it also indicates ``Rome (memorization) was not built in a day''.} that bypasses direct processing of the training data. Specifically, we select datasets categorized into three distinct types---context-independent, conventional, and factual---and redefine memorization as the ability to produce correct answers under these conditions. Our analysis then focuses on disparities between memorized and non-memorized samples by examining the logits and representations of generated texts.
Experimental findings reveal that longer words are less likely to be memorized, higher confidence correlates with greater memorization, and representations of the same concepts are more similar across different contexts.
\end{abstract}

\section{Introduction}
Memorization in the context of large language models (LLMs) commonly refers to the reproduction of text fragments from their training corpus  \cite{radford2019language,zhang2021counterfactual,tirumala2022memorization,carlini2023quantifying}. As foundational components in various NLP tasks, LLMs highlight the critical need to examine their propensity for memorization.  
Initially, the study for memorization is driven by data leakage, such as emails and health records,  from pre-trained corpora. This kind of data leakage poses privacy risks and increases susceptibility to attacks \cite{carlini2019secret,lee2022deduplicating}. Recent investigations aim to understand the extent of memorization, identify influencing factors, and devise strategies to minimize it. These studies show that LLMs have a significant capacity to memorize training data, influenced by factors like model size, context length, temperature, and duplication frequency \cite{carlini2023quantifying}. Deduplication in training corpora is an effective method to reduce memorization \cite{kandpal2022deduplicating}.

\begin{figure}[!t]
  \centering
  \includegraphics[width=0.95\linewidth]{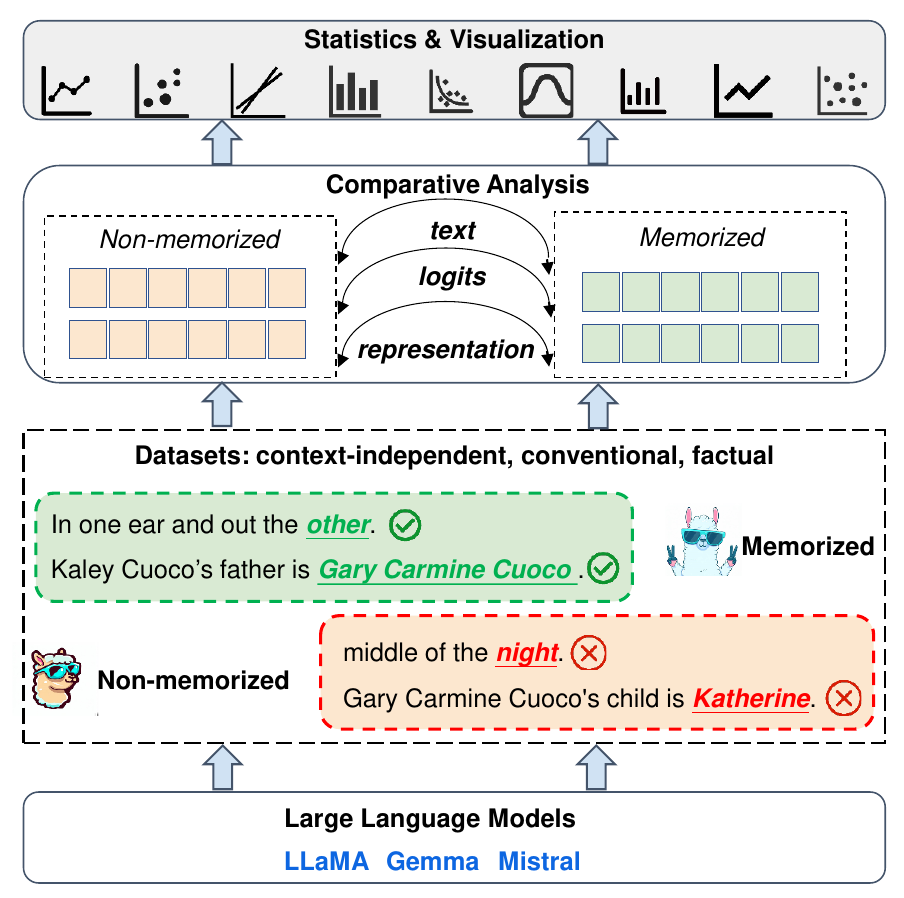}
  \caption{The framework of \modelname. 
  }
  \label{fig:intro}
  \vspace{-1.5 em}
\end{figure} 
Our study concentrates on investigating memorization characteristics in billion-scale LLMs, such as \llama~\cite{touvron2023llama}, \mistral~\cite{jiang2023mistral} and \gemma~\cite{team2024gemma}. Unlike existing work that quantifies memorization by directly comparing the model's outputs with its pre-training corpus, we employ  a different approach. Given the immense size of these corpora (e.g., 2 trillion tokens for \llama-2) and the intensive data processing requirements \cite{carlini2023quantifying,kandpal2023large}, we opt to bypass extensive corpus processing. 
As in previous studies, we employ factual question answering datasets to explore memorization. Additionally, we extend our investigation to data that can be input and output as a whole, which we categorize under the ``block in block out'' characteristic, see Section \ref{block_in_block_out} for detailed description for the characteristic.
Based on the ``lock in block out'' capability, we select datasets in two categories. The first category, which exhibits the strongest ``block in block out'' characteristics, includes idioms and Chinese poetry. These datasets are treated as standalone blocks that express fixed and context-independent semantics. The second category comprises datasets containing conventional concepts, such as proper nouns and terminologies, which are broadly defined and universally recognized.

Figure \ref{fig:intro} shows the framework of \modelname. A sample is categorized as \textit{memorized} if it is completed or answered correctly; otherwise, it is deemed \textit{non-memorized}. During this process, we also record logits and representations alongside the generated text. With memorized and non-memorized groups established, we conduct comparative analyses using statistical and visualization methods to unveil changes in the models' behavioral patterns.

The main contributions are outlined below:
\begin{itemize}
    \item We identify datasets that exhibit the ``block in block out'' characteristics and define ``memorized'' as  correctly answered or completed. This transformation helps us avoid processing and analyzing LLMs' training corpus.
    
    \item We explore memorization not only from the textual perspective but also through analysis of logits and representation, conducting comparative studies on these features between memorized and non-memorized samples.
    
    \item We present several empirical findings indicating that longer prompts are more likely to be memorized, whereas longer words are less likely; higher levels of memorization correlate with greater confidence and accuracy; and distinct representations exist between memorized and non-memorized samples, with similar concepts tending to exhibit more similar features.
\end{itemize}

\section{Related Work}
Memorization was first identified by \citet{radford2019language} in GPT-2 and has since been confirmed in other models such as \llama~and GPT-3.5 \cite{karamolegkou-etal-2023-copyright}. This presents significant privacy risks, as illustrated by \citet{carlini2021extracting}, who demonstrated GPT-2's ability to retrieve sensitive personal information including names, email addresses, phone numbers, and addresses. Such capabilities underscore the potential for privacy intrusions and attacks \cite{carlini2019secret, tirumala2022memorization, carlini2023quantifying}.

Originally, memorization was characterized by how well the generated sequences aligned with the actual continuations in the training data, employing different $n$-gram levels \cite{brown2020language,lee2022deduplicating,chen2021evaluating,ziegler2022first}. While these studies primarily address verbatim memorization, to capture more complex and semantically rich forms of repetition, \citet{feldman2020neural,van2021memorization,zhang2021counterfactual,chen2022decoupling} defined memorization as the performance variance of a training instance when included or excluded from the training set.

With these memorization definitions, various factors influencing memorization have identified.
\citet{zhang2021counterfactual} observed that the most memorized training examples are qualitatively distinct yet simple enough to be learned from a single instance. Additionally, \citet{chen2022decoupling,zheng2022empirical} noted that atypical samples often receive higher memorization scores. Exploring different dimensions, \citet{levy2021investigating} found that lowering a model's temperature and increasing its size enhances memorization of conspiracy theories, while \citet{kharitonov2021bpe} demonstrated the impact of subword vocabulary size on memorization.
Furthermore, \citet{tirumala2022memorization} showed  a tendency to memorize nouns and numbers first. \citet{carlini2023quantifying} established that memorization increases with model capacity, data duplication, and the length of the prompting context. Finally, \citet{haviv2023understanding} revealed that memorization in models is a two-step process where early layers promote a token to the top of the output distribution, and upper layers solidify this with increased confidence.
In addition, the negative impacts of memorization and strategies to mitigate it have also been explored, such as  \citet{lee2022deduplicating,kandpal2022deduplicating,hernandez2022scaling,carlini2023quantifying,biderman2023emergent} focused on reducing memorization by eliminating duplicate samples from training datasets. Their findings indicate that deduplication significantly lowers memorization, thereby enhancing security against privacy attacks.

Access to training datasets is often limited due to their non-public nature or large volume \cite{carlini2023quantifying}. For instance, generating 100 tokens per second on a V100 GPU, the 6B parameter GPT-Neo model would require over 30 GPU-years to process an 800GB training dataset \cite{carlini2023quantifying}. Consequently, some researchers have developed methods to investigate memorization without direct access to the training data. \citet{levy2021investigating} treated the model as a black box, analyzing outputs to infer memorization patterns without needing ground-truth data. Similarly, \citet{haviv2023understanding} constructed a separate dataset with known ground-truth to explore how memorization evolves across different layers.

Building upon these methodologies, our work advances the study of memorization. Similar to \citet{levy2021investigating,haviv2023understanding}, we also avoid directly accessing training data.  Instead, we use unique datasets with definitive answers, categorizing each sample into \textit{memorized} and \textit{non-memorized} based on whether it is answered correctly. Rather than focusing on the factors that influence memorization, we delve into the model's behavioral patterns during memorization by examining the differences between these two categories.

\section{Methodology}
The following contents will detail the definition of memorization, describe the ``block in block out'' characteristics, outline the selected datasets, and discuss the analytical perspectives employed in this study.

\paragraph{Definition of memorization.}
Considering a text segment $S$ and its context $C$, we represent this as 
{\small\begin{equation}
S+C = \{w_1, \cdots, w_{k-1}, \underline{w_k, \cdots, w_i, w_n}, w_{n+1}, \cdots, w_m\}, 
\end{equation}}
where $S = \{w_k, \cdots, w_i,  w_n \}$ denotes the text segment, $C_l = \{w_1, \cdots, w_{k-1} \}$ and $C_r = \{w_{n+1}, \cdots, w_m \}$ are its contexts. 
For question-answering problem, the sample is categorized as \textit{memorized} if the model answers it correctly. For complete task, given $S' = \{w_k, \cdots, \_, w_n \}$, if the model can correctly predict $w_i$ for $(k \leq i \leq n)$, the sample is categorized as \textit{memorized}; otherwise, it is \textit{non-memorized}.

\paragraph{Block in block out.} \label{block_in_block_out}
This definition of memorization requires the datasets to exhibit a characteristic where the whole is greater than the sum of its parts. This trait emphasizes that the meaning of a text segment is derived from the collective function of its entire structure, rather than merely from an aggregation of individual elements. For instance, the idiom ``hit the nail on the head'' and the Chinese poem ``\zhsmall{千山鸟飞绝} (from hill to hill no bird in flight)'' both exemplify this feature. Isolating any single word from these expressions fails to convey the full meaning. Text segments that satisfy this characteristic are referred to as capable of ``block in block out'', meaning they are input to pre-train the LLMs, and the models must reproduce them as a whole.  For a text segment $S$, if a new text $S'$ is obtained by addition, deletion, or modification, the rarer $S'$ occurs and the more significant its impact on the semantics of $S$, the higher the likelihood that $S$ serves as a ``block in block out'' instance. We describe the concept of ``block in block out'' rather than quantifying it.

\subsection{Datasets}\label{sec:dataset}
In order to be able to ``block in block out'', we have selected two categories of datasets. Additionally, we include three factual question-answering datasets. Please refer to Appendix \ref{sec:examples} for examples of each dataset.

\paragraph{Context-independent.} This category includes idiom and Chinese poetry datasets, which are particularly suited for the ``block in block out'' approach. The semantics of these datasets are usually pertinent to a specific culture and feature definitive answers that remain invariant across different contexts and models, thus making them context-independent.

\textbf{\datasetidioms} \cite{haviv2023understanding} is a collection of English idioms (e.g., no pain no gain) designed to assess memorization . It comprises 850 samples, with an average of 4.9 words per sample.

\textbf{\datasettp} comprises ancient Chinese poems collected from the ``Three Hundred Tang Poems'' a poetry anthology from the Tang Dynasty (e.g., \zhsmall{千山鸟飞绝} (from hill to hill no bird in flight)). We have selectively retained the main body of five-character quatrains and seven-character quatrains, ultimately obtaining approximately 1500 samples.

\paragraph{Conventional.} The second category includes datasets of proper nouns and disease terminologies.  These datasets are typically certified by authoritative institutions to reduce ambiguities in communication, ensuring that each concept or entity is consistently associated with a unique text fragment. However, these concepts or entities may be represented by multiple different texts before receiving official certification. As a result, their ``block in block out'' characteristic is generally weaker compared to the datasets of the first category.

\textbf{\datasetpn} comprises a collection of proper nouns (e.g., Royal College of Art), generated with the assistance of artificial intelligence tools. We tasked \gptfour~to retrieve 200 rare and 200 common proper nouns, each consisting of four or more words. We then manually linked the retrieved nouns to their corresponding entries on the Wikipedia website, retaining only those that have a Wikipedia presence. We ultimately obtained approximately 300 proper nouns.

\textbf{\datasetterm} was also generated from \gptfour~consisting of disease terminologies (e.g., Acquired Immunodeficiency Syndrome). We requested 200 common and 200 rare disease terms, each comprising three or more words. We then manually verified these terminologies against The International Statistical Classification of Diseases and Related Health Problems 10th Revision (ICD-10), retaining only those terms that were assigned ICD-10 codes. We ultimately retained about 200 terminologies.

\begin{table}[!t]
\centering
\resizebox{0.9\columnwidth}{!}{
\begin{tabular}{lcc}
\toprule
&   \textbf{GPT-3.5} &    \textbf{\llama-3 8B} \\
\midrule
Total&  1513&   1513    \\
\hline
Correctly answer parents&   755&    554\\
Correctly answer child\\(\textit{w/} context) &    673& 441\\
Correctly answer child\\(\textit{w/o} context) &    248& 106\\
\bottomrule
\end{tabular}
}
\caption{Correct answers to ``Name a Child of Someone'' with/without a preceding context.}
\label{tab:mem_reason}
\end{table}

\paragraph{Factual.} In addition to the datasets specifically designed for ``block in block out'' characteristics, we also test several factual question answering datasets previously used to study memorization \cite{haviv2023understanding}, such as \datasetpopqa, \datasetlama~and \datasetcele.

\textbf{\datasetpopqa} \cite{mallen2023not} is an English question answering dataset, consisting of questions about someone's occupation, birthplace, genre, and capital, among others. It is created by converting knowledge tuples retrieved from Wikidata using templates and contains about 600 questions.

\textbf{\datasetlama} \cite{kassner2021multilingual} is similar to \datasetpopqa, derived from triples of the form (object, relation, topic), but specifically restricted to location-based answers. It is a subset of LAMA \cite{petroni2019language}, from which easily guessable examples have been filtered out. It consists of about 500 samples.

\textbf{\datasetcele} \cite{berglund2023reversal} is an English reversal relation dataset, for example, asking ``Who is Tom Cruise's mother?'' and its inverse ``Name a child of Mary Lee Pfeiffer''. It contains a list of the top 1000 most popular celebrities from IMDB (2023) and their parents.  This dataset was chosen because the same entity in question and answer remains consistent while only their context changes, allowing us to control ``frequency'', a key factor influencing memorization \cite{carlini2023quantifying}. The dataset includes about 1500 child-parent pairs.

\begin{figure*}[!t]
  \centering
    \begin{subfigure}{0.195\textwidth}
    \includegraphics[width=\linewidth]{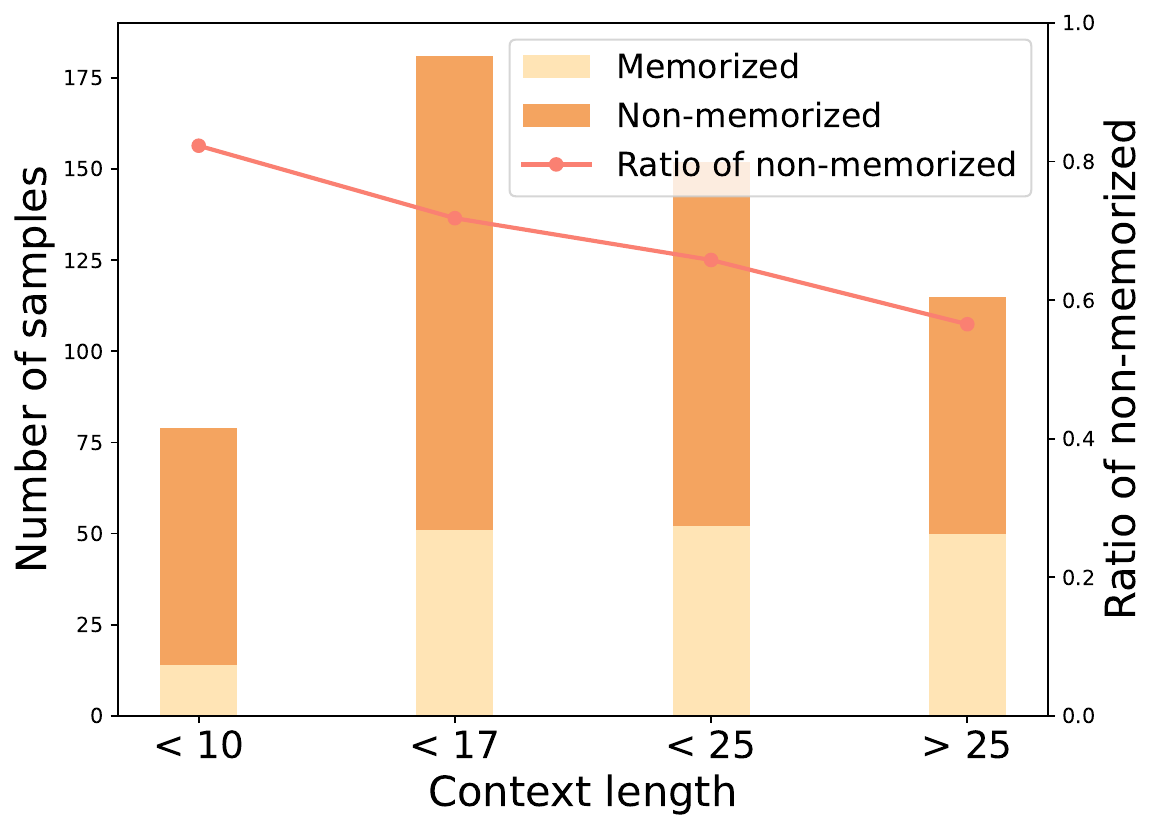}
  \end{subfigure}
  \hfill
  \begin{subfigure}{0.195\textwidth}
    \includegraphics[width=\linewidth]{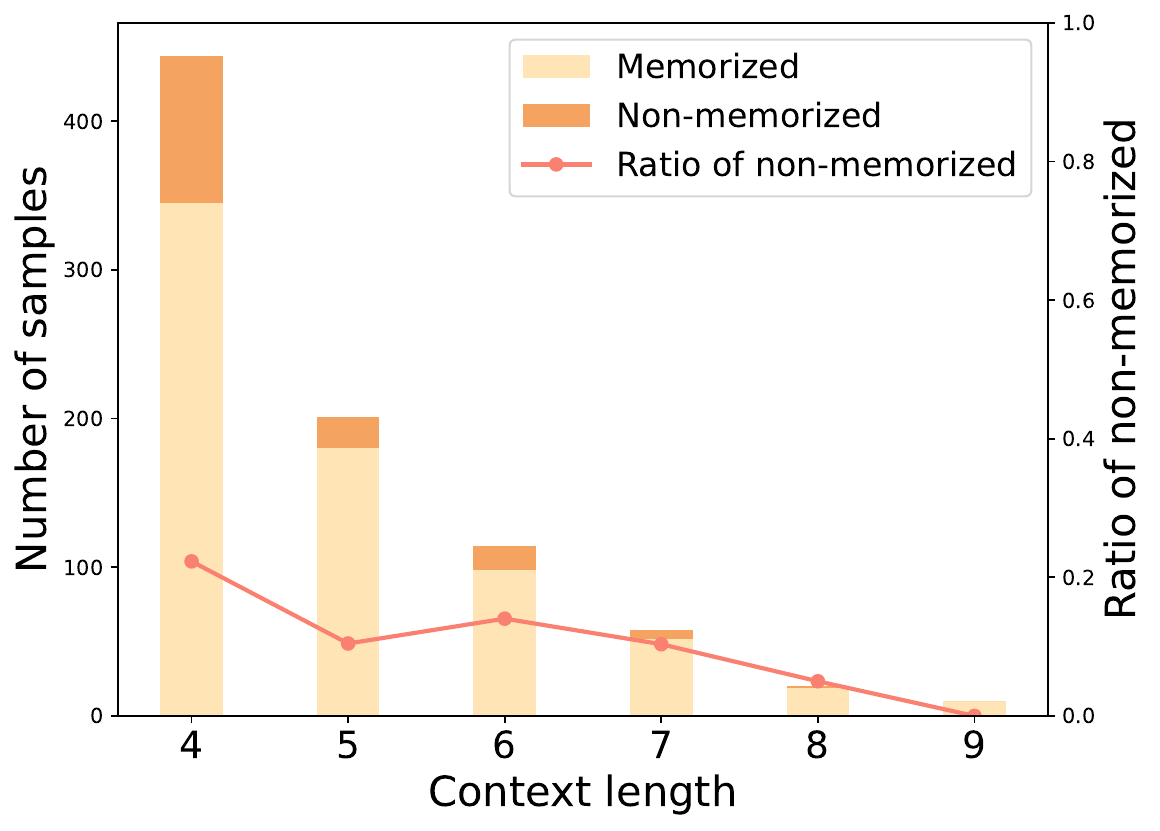}
  \end{subfigure}
  \hfill
  \begin{subfigure}{0.195\textwidth}
    \includegraphics[width=\linewidth]{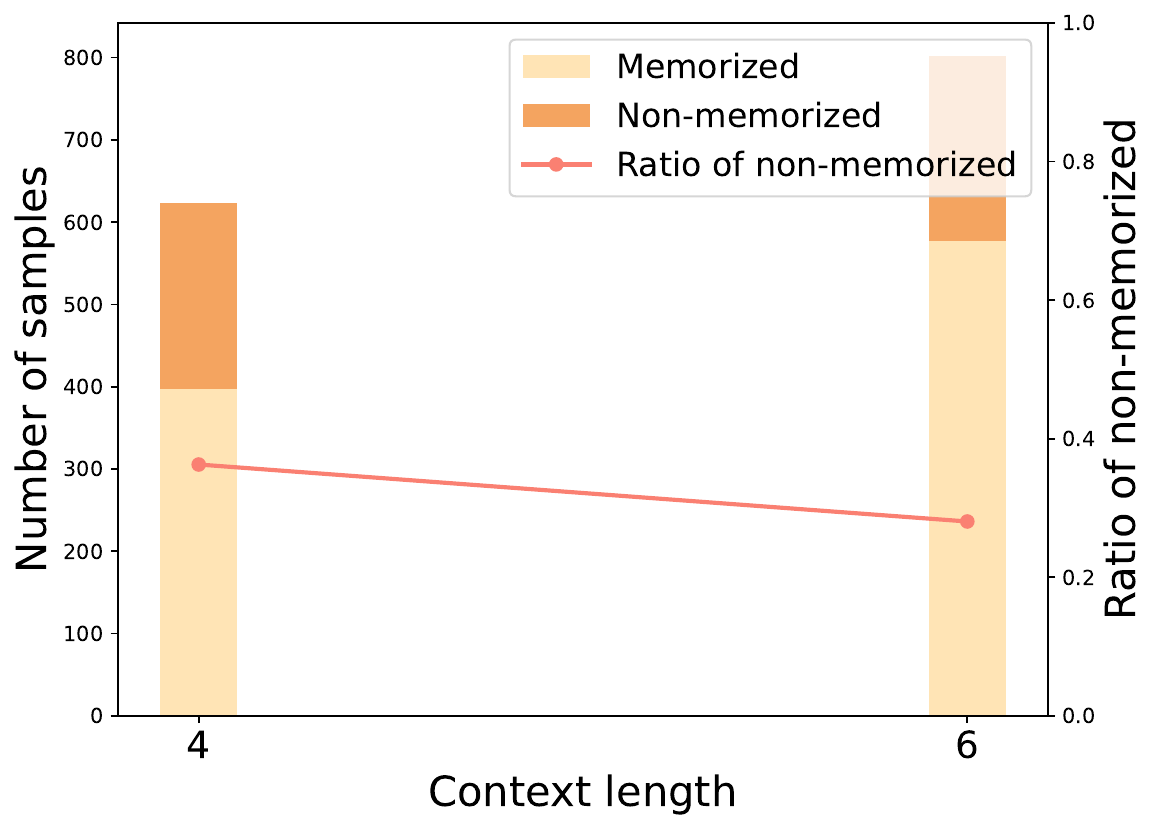}
  \end{subfigure}
  \hfill
  \begin{subfigure}{0.195\textwidth}
    \includegraphics[width=\linewidth]{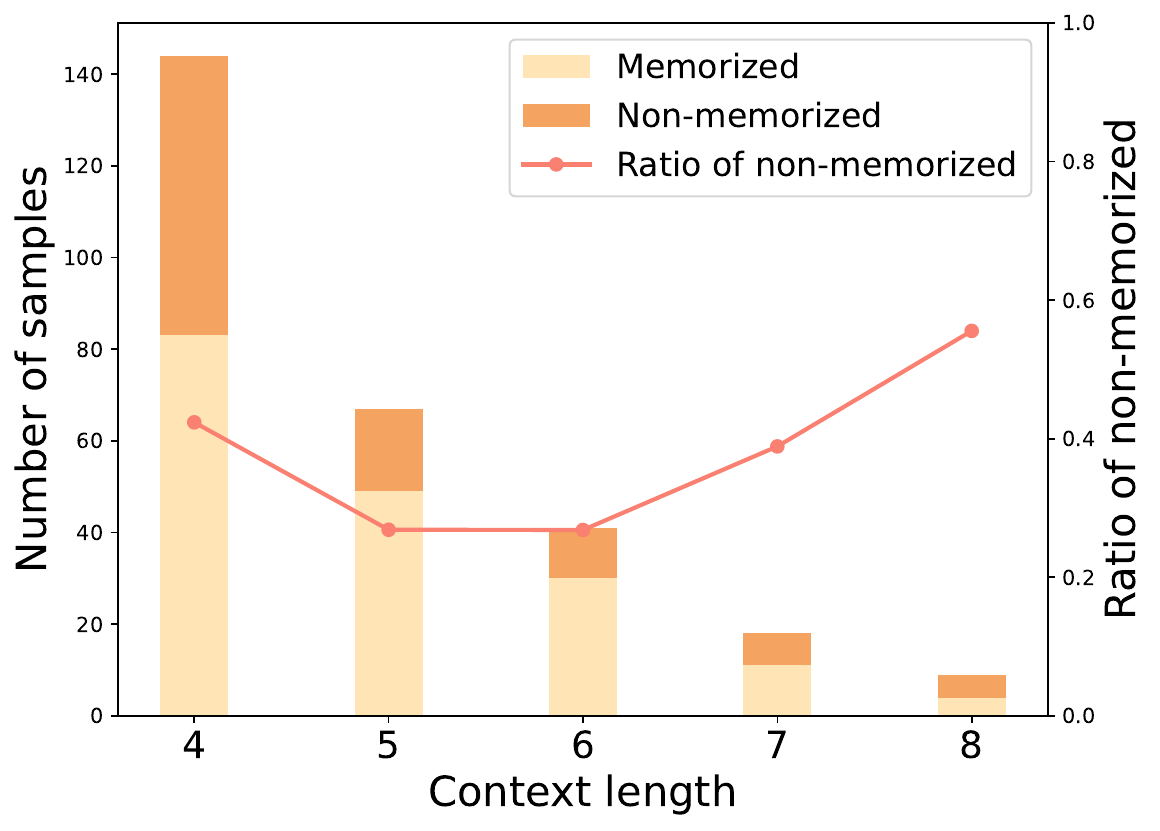}
  \end{subfigure}
  \hfill
  \begin{subfigure}{0.195\textwidth}
    \includegraphics[width=\linewidth]{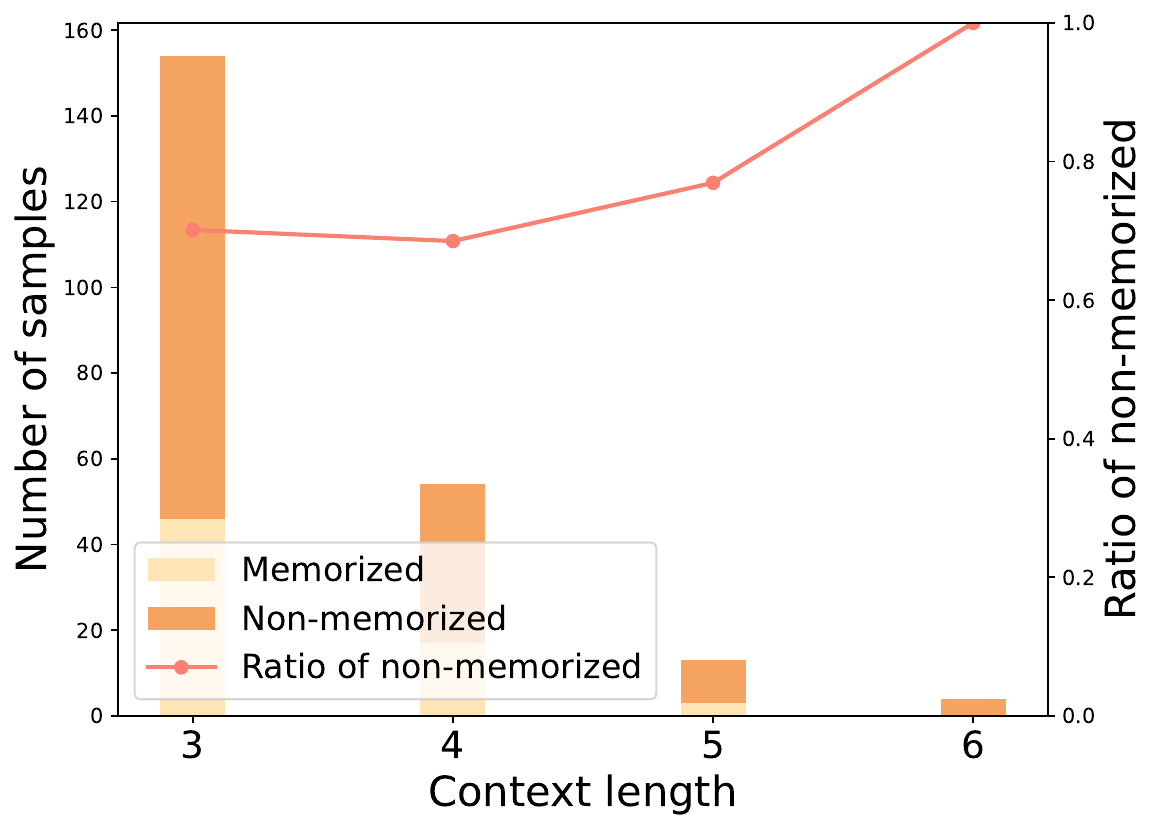}
  \end{subfigure}
    \begin{subfigure}{0.195\textwidth}
    \includegraphics[width=\linewidth]{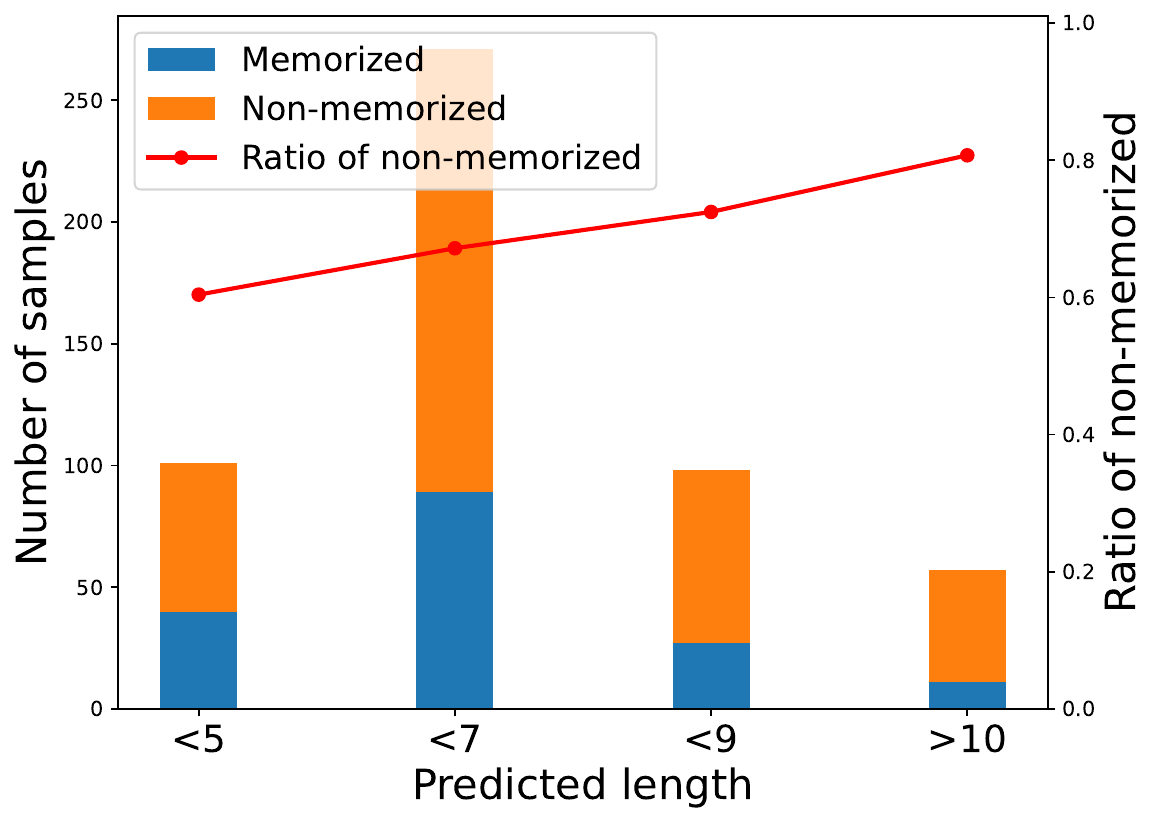}
    \caption{\datasetlama}
  \end{subfigure}
  \hfill
  \begin{subfigure}{0.195\textwidth}
    \includegraphics[width=\linewidth]{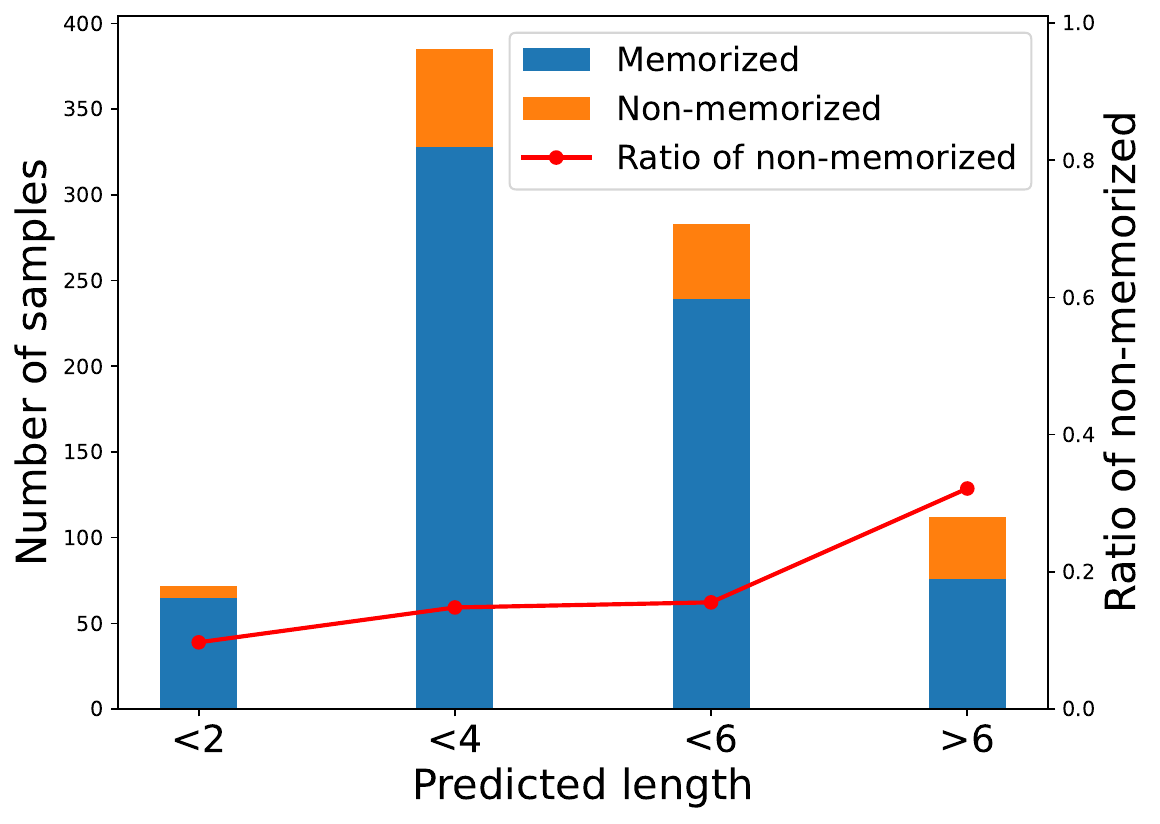}
    \caption{\datasetidioms}
  \end{subfigure}
  \hfill
  \begin{subfigure}{0.195\textwidth}
    \includegraphics[width=\linewidth]{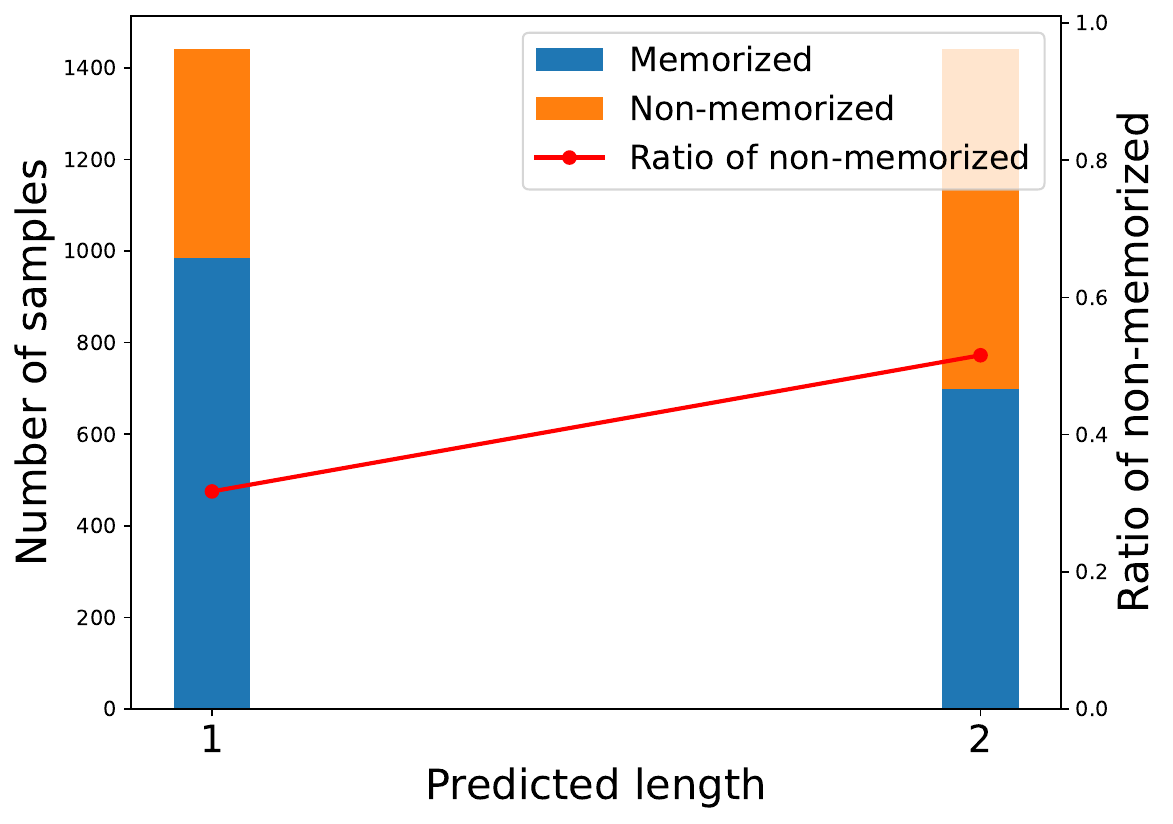}
    \caption{\datasettp}
  \end{subfigure}
  \hfill
  \begin{subfigure}{0.195\textwidth}
    \includegraphics[width=\linewidth]{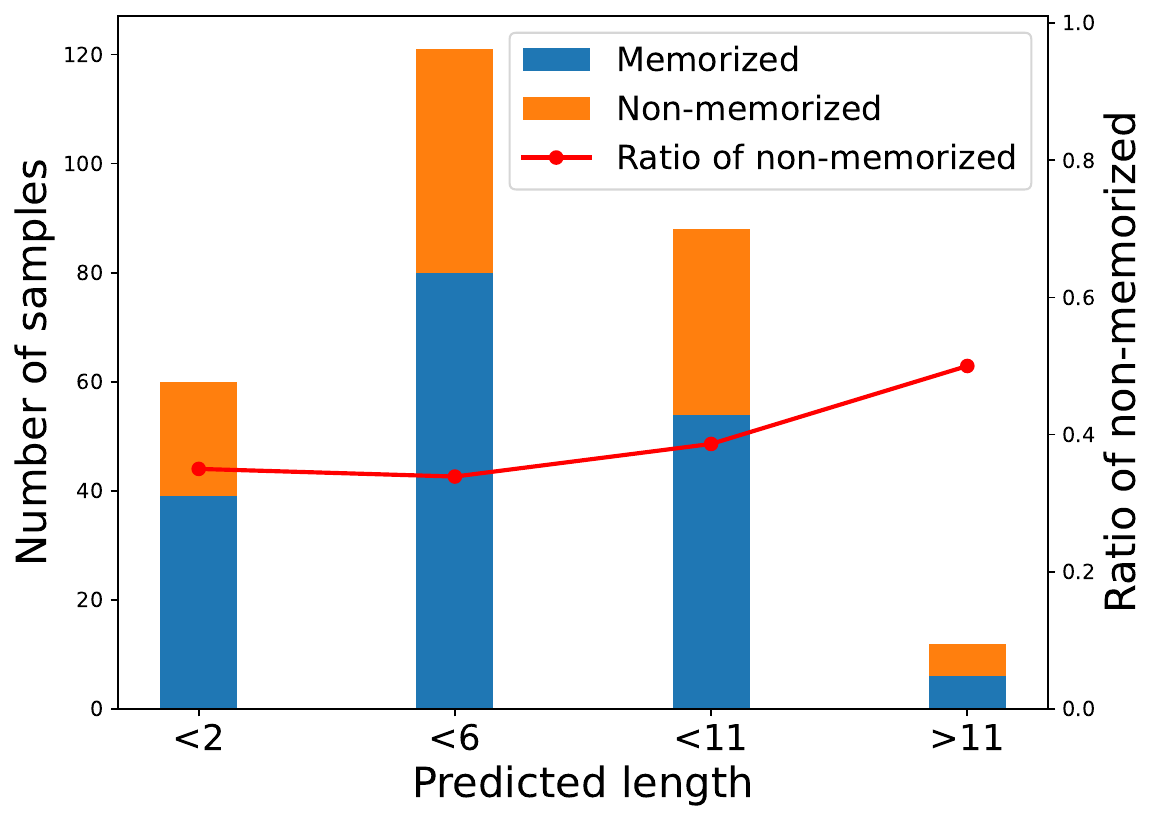}
    \caption{\datasetpn}
  \end{subfigure}
  \hfill
  \begin{subfigure}{0.195\textwidth}
    \includegraphics[width=\linewidth]{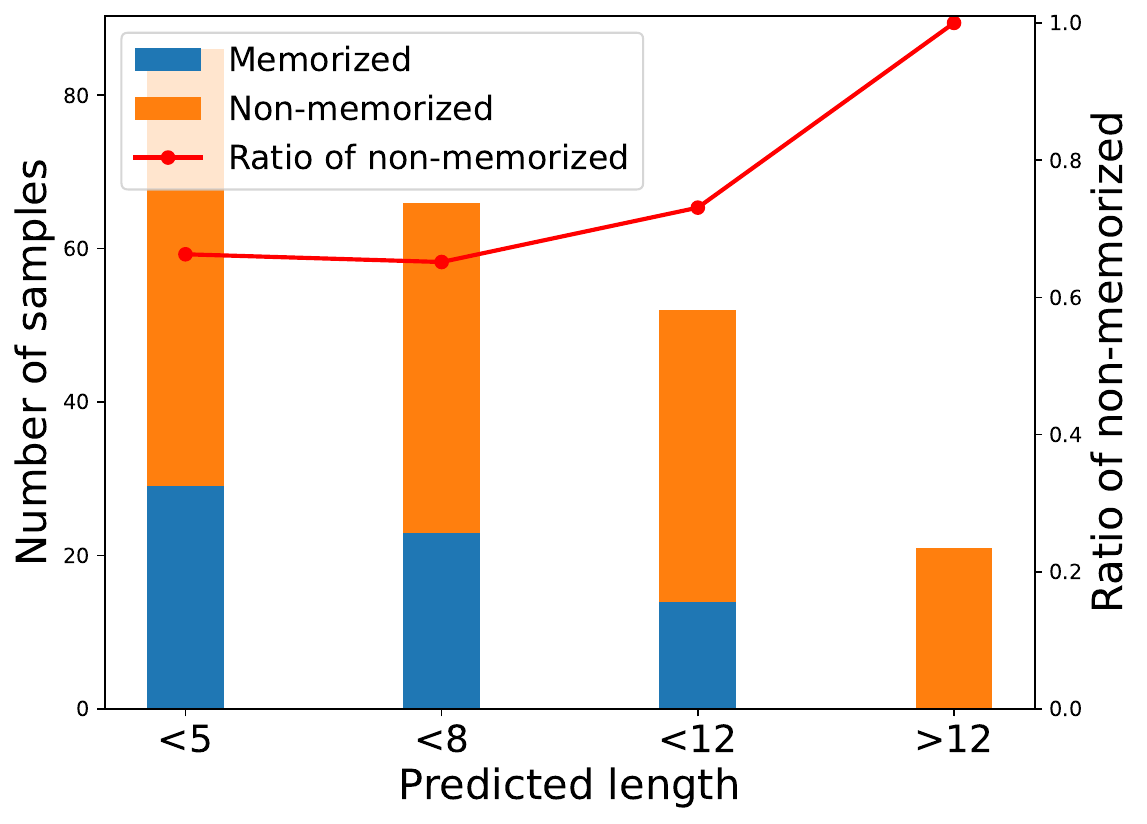}
    \caption{\datasetterm}
  \end{subfigure}
  \caption{Comparison of memorized vs non-memorized instances: context and predicted length across datasets (\datasetlama, \datasetidioms, \datasettp, \datasetpn, \datasetterm).}
  \label{fig:text_results}
  \vspace{-1 em}
\end{figure*}

\paragraph{Memorization or reasoning?} Some previous works also suggest that factual question-answering tasks may primarily stimulate a model's reasoning abilities \cite{wang2024chain}. However, we tend to believe that these tasks invoke memorization capabilities. To support our hypothesis, we conducted experiments on \datasetcele~to determine whether it leans more towards memorization or reasoning. We tested the model's ability to answer ``Name a Child of someone'' with or without context. Without context, the question was posed directly; with context, it was preceded by ``Who is someone's parents?''. After receiving the model's response, ``Name a child of someone'' was asked. We analyzed the correctness of the model's responses, which are shown in Table  \ref{tab:mem_reason}.
It can be observed that when there is a preceding question (i.e., w/ context), such as ``Who is Tom Cruise’s mother?'', the accuracy rates of GPT-3.5 and \llama-3 8B in answering subsequent questions, such as ``Name a child of Mary Lee Pfeiffer'', are 44.5\% and 29.1\%, respectively. However, without the preceding question (i.e., w/o context), the accuracy rates for both models drop dramatically to 16.4\% and 7.1\%. If such questions were primarily stimulating the model’s reasoning capabilities, the results of w/ or w/o context should be similar. Because, if the models can correctly answer questions like ``Who is Tom Cruise’s mother?'', they should also be able to infer the answers to questions such as ``Name a child of Mary Lee Pfeiffer'' from the knowledge they have previously learned. The substantial discrepancy observed suggests that answering the question ``Name a Child of someone'' primarily invokes the model’s memorization abilities. For w/o context, not only is it simpler for the model to answer about a celebrity’s parents, but it can also infer the parent-child relationship from the context provided, thus GPT-3.5 and \llama-3 8B increase the accuracy of answering the questions from (16.4\%, 7\%) to (44.5\%, 29.1\%), respectively.

\subsection{Insights}
As is widely recognized, in the generation process of GPT-style generative models, the model first calculates a representation based on previous tokens. It then uses this representation to compute the distribution over the vocabulary (i.e., logits). The next token is subsequently generated using decoding strategies \cite{radford2018improving,radford2019language,brown2020language}.  Therefore, in addition to analyzing the most commonly used {text}, we also examine the {logits} and representations.

\paragraph{Text.}
Attributes associated with text, such as word frequency, context length, and part-of-speech, have been extensively explored. To align with previous research, we re-examine the context length and predicted length under our defined memorization. Context length refers to the number of words in a context (excluding exemplars), while predicted length refers to the number of characters in the word to be predicted.

\paragraph{Logits \& Representation.}
If {text} represents the returned results, then {logits} can be considered the basis for this result's decision-making. Analyzing it enables us to comprehend the model's generative biases in \textit{memorized} versus \textit{non-memorized}. The {representation} of next token carries the model's understanding of the current context, and analyzing it can further enhance our understanding of the model's behavior.

\begin{table*}[!t]
\centering
\large
{\renewcommand{\arraystretch}{1.6}
\resizebox{\textwidth}{!}{
\begin{tabular}{l|l|cc|cc|ccc}
\toprule
\multirow{2}{*}{\textbf{Models}} & \multirow{2}{*}{\textbf{Datasets}} & \multicolumn{2}{c|}{\textbf{Context-independent}} & \multicolumn{2}{c|}{\textbf{Conventional}} & \multicolumn{3}{c}{\textbf{Factual}} \\ 
\cline{3-9} 
 &  & \textbf{\datasetidioms} & \textbf{\datasettp} & \textbf{\datasetpn} & \textbf{\datasetterm} & \textbf{\datasetcele} & \textbf{\datasetpopqa} & \textbf{\datasetlama} \\ 
 \toprule
\multirow{2}{*}{\mistral~7B} & \textit{memorized} & \textbf{0.5777} & -- & \textbf{0.5833} & \textbf{0.6669} & \textbf{0.8446} & \textbf{0.8343} & \textbf{0.6938} \\
 & \textit{non-memorized} & 0.5373 & -- & 0.4918 & 0.5891 & 0.6184 & 0.5895 & 0.4954 \\
 \midrule
\multirow{2}{*}{\gemma~7B} & \textit{memorized} & \textbf{0.4051} & -- & \textbf{0.7502} & \textbf{0.6768} & \textbf{0.8194} & \textbf{0.8134} & \textbf{0.6655} \\
 & \textit{non-memorized} & 0.2570 & -- & 0.6528 & 0.5435 & 0.6616 & 0.5418 & 0.5391 \\
 \midrule
{\llama-2 7B/} & \textit{memorized} & \textbf{0.3968} & \textbf{0.8206} & \textbf{0.6314} & \textbf{0.6735} & \textbf{0.8568} & \textbf{0.8140} & \textbf{0.6588} \\
 \qwen-1.5 7B& \textit{non-memorized} & 0.3098 & 0.3072 & 0.3298 & 0.5149 & 0.6502 & 0.6136 & 0.3369 \\
 \midrule
{\llama-2 13B/} & \textit{memorized} & \textbf{0.4254} & \textbf{0.8376} & \textbf{0.7381} & \textbf{0.7646} & \textbf{0.8371} & \textbf{0.8362} & \textbf{0.6789} \\
 \qwen-1.5~14B& \textit{non-memorized} & 0.3279 & 0.3180 & 0.3780 & 0.5489 & 0.6787 & 0.5386 & 0.4759 \\
 \midrule
{\llama-3 8B/} & \textit{memorized} & \textbf{0.3607} & \textbf{0.8760} & \textbf{0.746} & \textbf{0.6917} & \textbf{0.8131} & \textbf{0.8295} & \textbf{0.6770} \\
 \qwen-1.5 32B& \textit{non-memorized} & 0.2816 & 0.3876 & 0.5421 & 0.6200 & 0.6835 & 0.5725 & 0.5693 \\
 \bottomrule
\end{tabular}
}
}
\caption{Comparison of averaged probability of generated tokens between \textit{memorized} and \textit{non-memorized}.}
\label{tab:probability_analysis}
\end{table*}

\section{Experimental Analysis}
\paragraph{Parameter Settings.}
We conduct experiments using various model configurations including \llama-2 (7B, 13B), \mistral~7B, \gemma~7B, and \llama-3 8B. These models are loaded onto two V100 GPUs, each with 32GB of memory, and operated in float32 precision. During the decoding phase, we employ a greedy decoding strategy \cite{germann-2003-greedy} that selects the token with the highest probability.

By default, given a text of length $n$, for \datasetidioms~and \datasettp, we utilize the first $n-1$ items as prompts and designate the last word as the gold text. For \datasetpn~and \datasetterm, the penultimate word is used as the gold text, while the remaining words serve as prompts. For \datasetpopqa, \datasetlama~and \datasetcele, the original question is directly employed as the prompt. For \datasetcele, the maximum number of generated tokens is set to 8. For the Chinese \datasettp~dataset, the maximum number of tokens produced is 1, and for all other datasets, the limit is set at 4 tokens. Finally, we categorize the samples as memorized or non-memorized based on exact match criteria. For predicted text that comprises multiple tokens, we average their probabilities (the logits normalized through the softmax function) or representation. 
For the exemplars used for each dataset, please refer to Appendix \ref{sec:exemplars}.

\subsection{Text-oriented Analysis.}\label{sec:text_oriented_analysis}
To investigate whether there are distinct disparities between memorized and non-memorized samples concerning context length and predicted length, we analyze the number and proportion of memorized versus non-memorized samples. Given the consistency of experimental results across different models, we present data only from \llama-2 13B for English datasets and \qwen-1.5 \cite{bai2023qwen} 32B for the Chinese dataset, as shown in Figure \ref{fig:text_results}.

\paragraph{Longer prompt more memorized?} The first analysis examines the relationship between context length and the likelihood of samples being memorized. Consistent with previous studies, we find that the proportion of non-memorized samples decreases as context length increases in datasets such as \datasetlama, \datasetidioms, and \datasettp. In contrast, \datasetpn~and \datasetterm~exhibit a different pattern: the ratio of non-memorized samples first decreases and then increases with context length. For \datasetidioms~and \datasettp, their text blocks, which convey context-independent semantics as wholes, benefit from longer contexts which provide more prior information and simplify the model’s task of recalling memorized content. However, for \datasetpn~and \datasetterm, which also express proprietary concepts or entities as wholes, the semantic combination of individual words within the text still retains original meaning to some extent. Thus, overly short or long contexts may increase the difficulty of recall due to the increased internal diversity of potential substitute words.

\paragraph{Longer word less memorized.} The second analysis explores the relationship between predicted length and the likelihood of memorization. We observe a linear trend where, as the number of characters in a word increases, the ratio of non-memorized samples also continuously increases. This pattern likely arises from the increased complexity associated with longer words, which results in lower memorization rates.

\begin{figure}[!t]
    \centering
    \begin{subfigure}{0.49\columnwidth}
        \includegraphics[width=\linewidth]{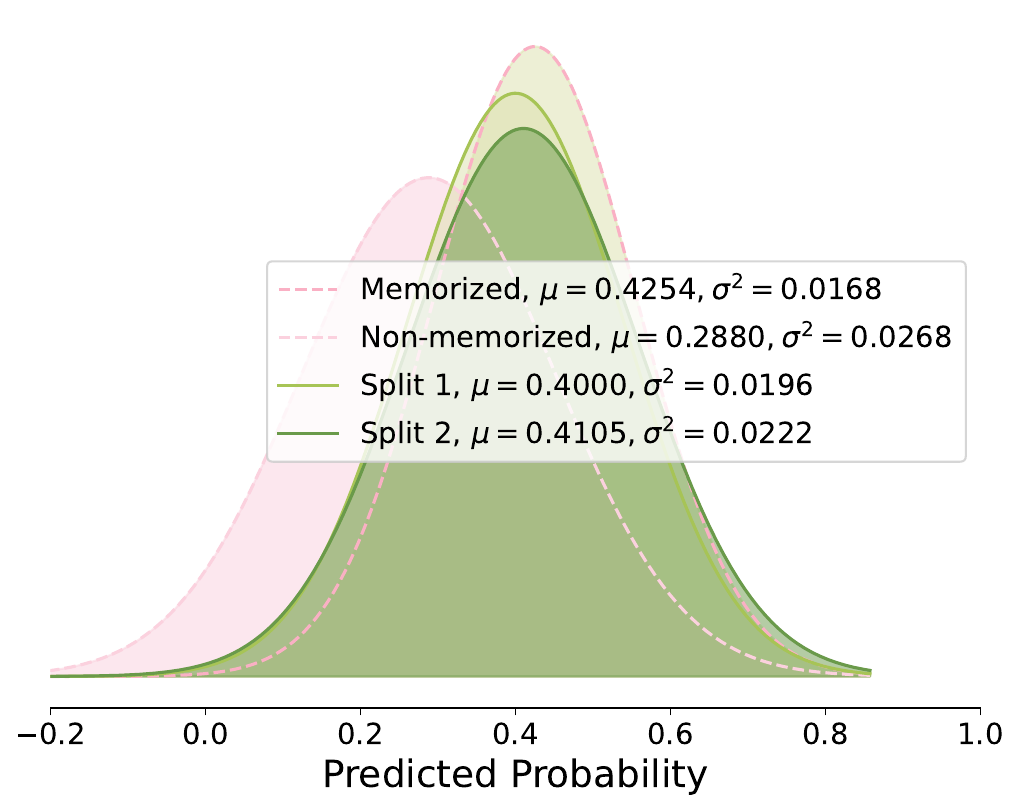}
        \caption{\datasetidioms}
    \end{subfigure}
    \hfill
    \begin{subfigure}{0.49\columnwidth}
        \includegraphics[width=\linewidth]{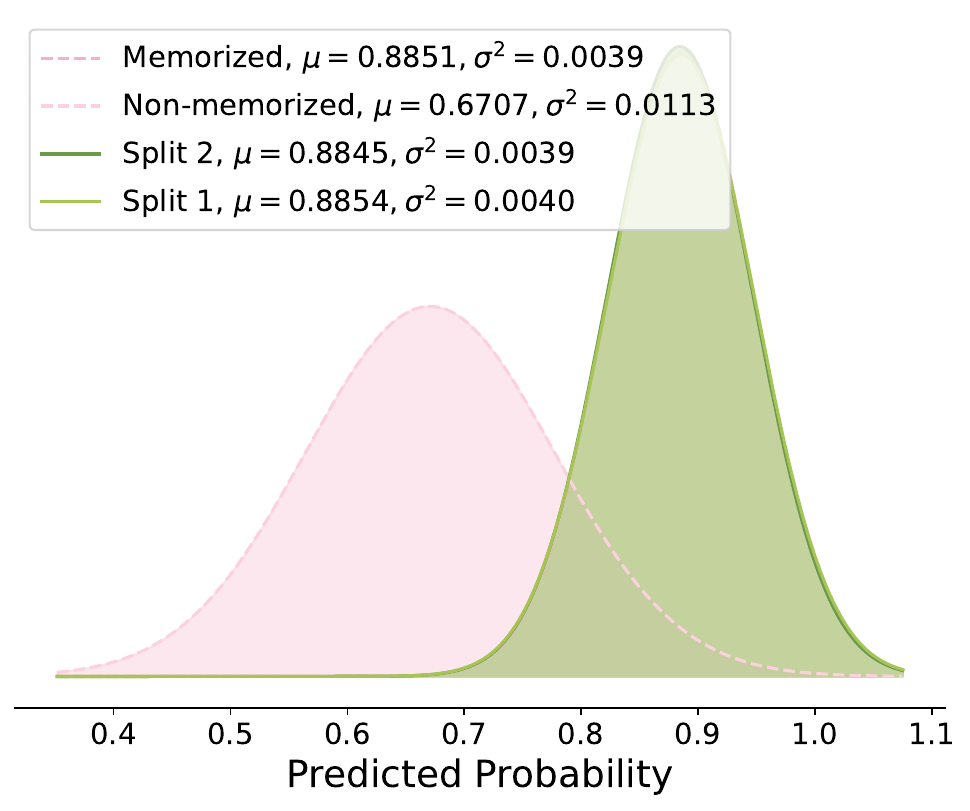}
        \caption{\datasetcele}
    \end{subfigure}
\caption{\textit{Probability} statistics across random splits.}
\label{fig:random_probability}
\end{figure}

\subsection{Logits-oriented Analysis.}\label{sec:probability_oriented_analysis}
To examine whether there is a significant change in the distribution of logits for the next token generated by the model between memorized and non-memorized samples, we record the normalized values of logits, i.e., the probability for each generated token, as shown in Table \ref{tab:probability_analysis}. For Chinese dataset (i.e., \datasettp) \qwen-1.5 models are used.

\paragraph{Higher memorized greater confidence.} 
We notice a marked disparity between the memorized and non-memorized groups, wherein memorized samples typically  exhibit higher probability scores. For instance, in the \datasetpopqa, the \llama-2 13B model shows average probabilities of 0.8362 for memorized instances compared to 0.5386 for non-memorized instances. This suggests that the model displays higher confidence in its memorized predictions.
To mitigate the impact of randomness on experimental outcomes, we randomly divided the dataset samples into two groups and computed the mean and variance of the probability values for these groups.  As illustrated in Figure \ref{fig:random_probability}, it displays the results for the \datasetidioms~and \datasetcele~using \llama-2 13B model. Given the consistency of these results across other datasets and models, and due to space limitations, only selective results are presented.   It is evident from the visualization that there is no significant difference in the probability distributions between the two randomly split groups, confirming the robustness of the findings.

\begin{figure}[!t]
\centering
    \includegraphics[width=0.9\columnwidth]{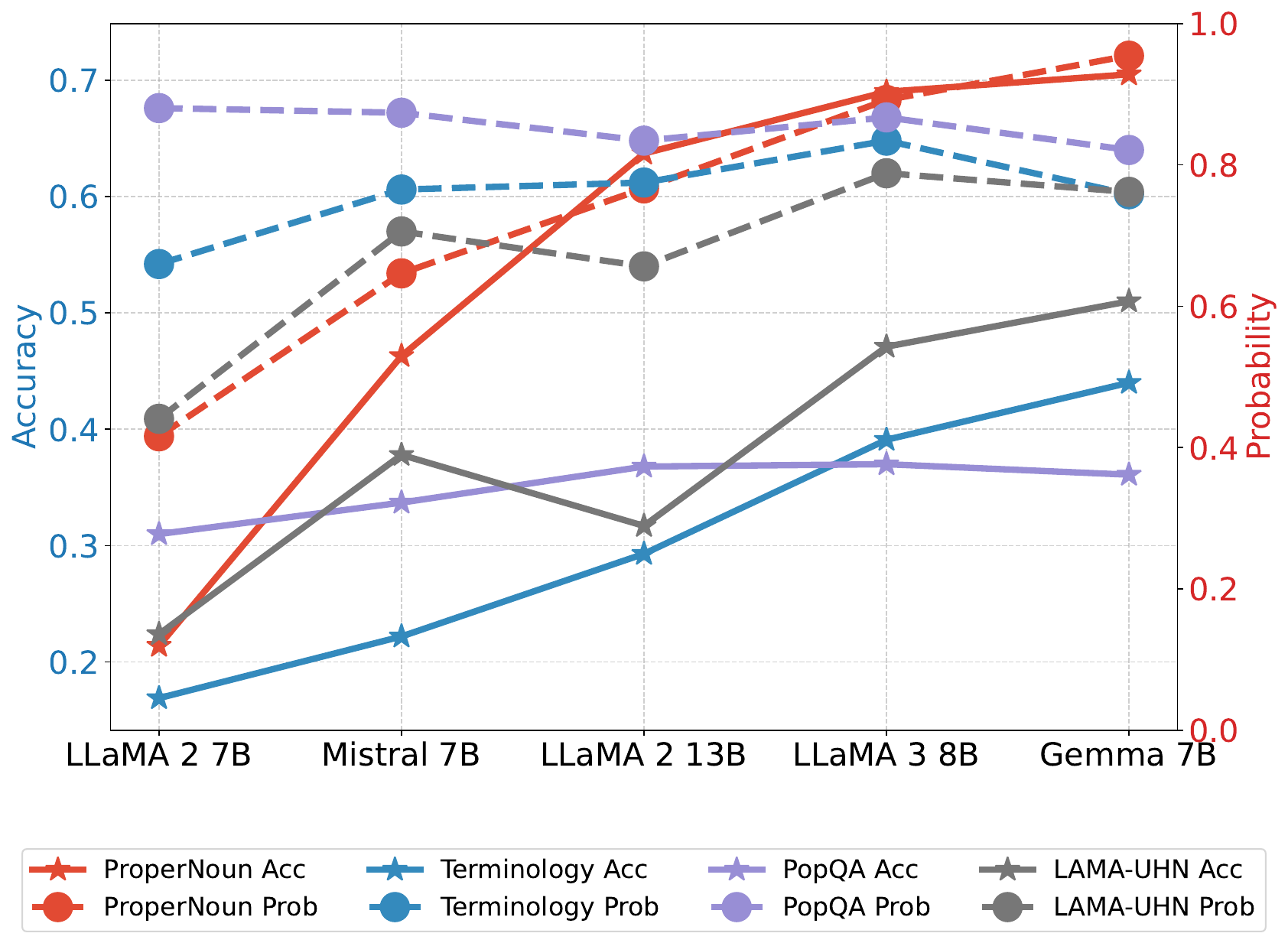}
\caption{Comparison between \textit{probability} and accuracy across all tested models and datasets.}
\label{fig:model_logits}
\end{figure}

\paragraph{Higher probability higher accuracy.}
Table \ref{tab:probability_analysis} illustrates that the normalized logits values for memorized samples are significantly higher. Based on this observation, we investigate whether higher probability values correlate with increased accuracy in generation. In other words, we explore if there is a consistent trend between model performance and probability. To this end, we record both the accuracy and probability values of all tested models across various datasets, as illustrated in Figure \ref{fig:model_logits}. Results are consistent across all English datasets; however, for clarity, only four datasets are presented here.

The data reveal a consistent correlation between accuracy and probability, suggesting that as accuracy increases, so does the probability, and vice versa. Among the five models evaluated, the \gemma~7B model demonstrates superior performance across all datasets, with an average accuracy of 0.564 and an average probability of 0.589. In contrast, the \llama-2 7B model shows generally poorer performance, with an average accuracy of 0.347 and an average probability of 0.480. Interestingly, despite having the largest parameter count, the \llama-2 13B model only exhibits moderate performance but still outperforms the \llama-2 7B model released concurrently.

Further analysis reveals that as probability values increase, there is a tendency for them to cluster, indicating a concentration of higher probability values. Conversely, as probability values decrease, they tend to diverge, suggesting that lower probability values are more dispersed across the datasets. This pattern further supports the notion that higher probabilities are associated with more consistent and accurate model predictions.

\begin{figure*}[!t]
  \centering
  \includegraphics[width=0.16\linewidth]{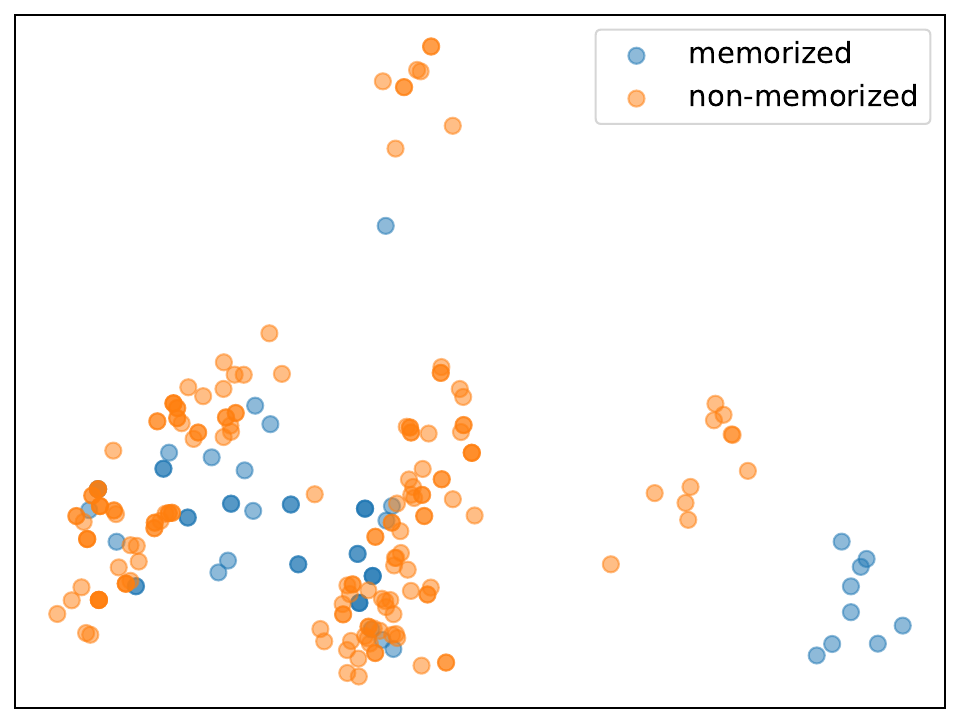}
  \includegraphics[width=0.16\linewidth]{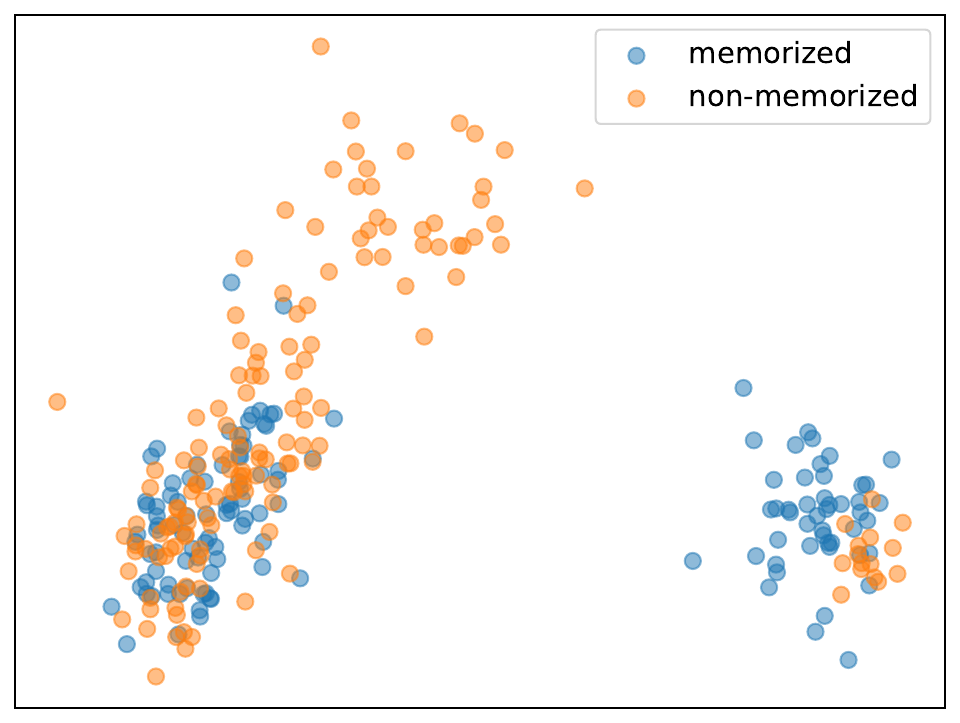}
  \includegraphics[width=0.16\linewidth]{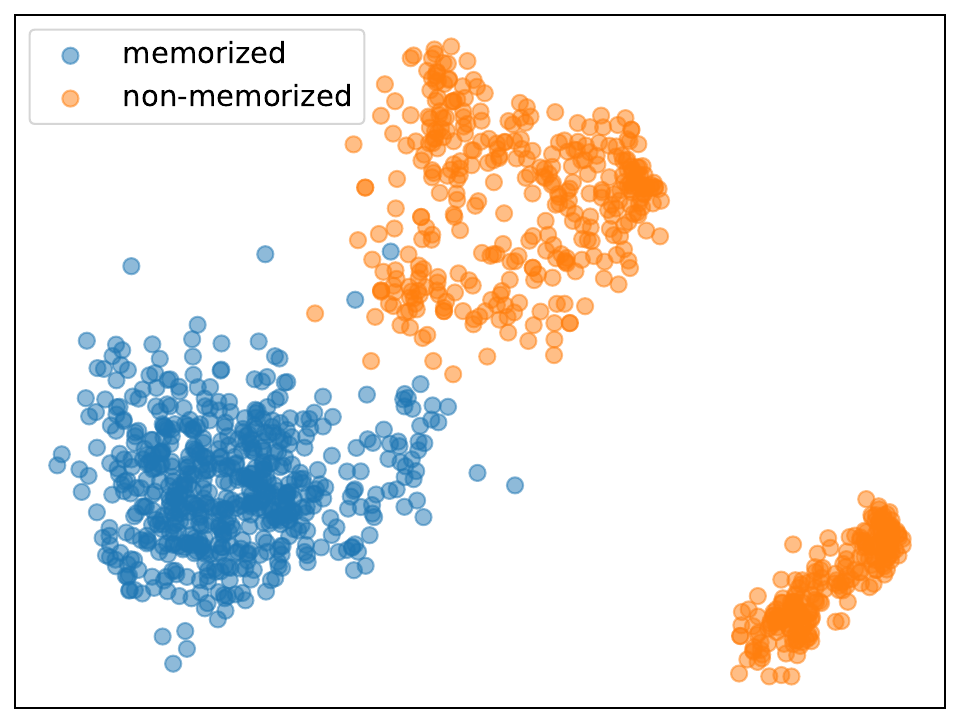}
  \includegraphics[width=0.16\linewidth]{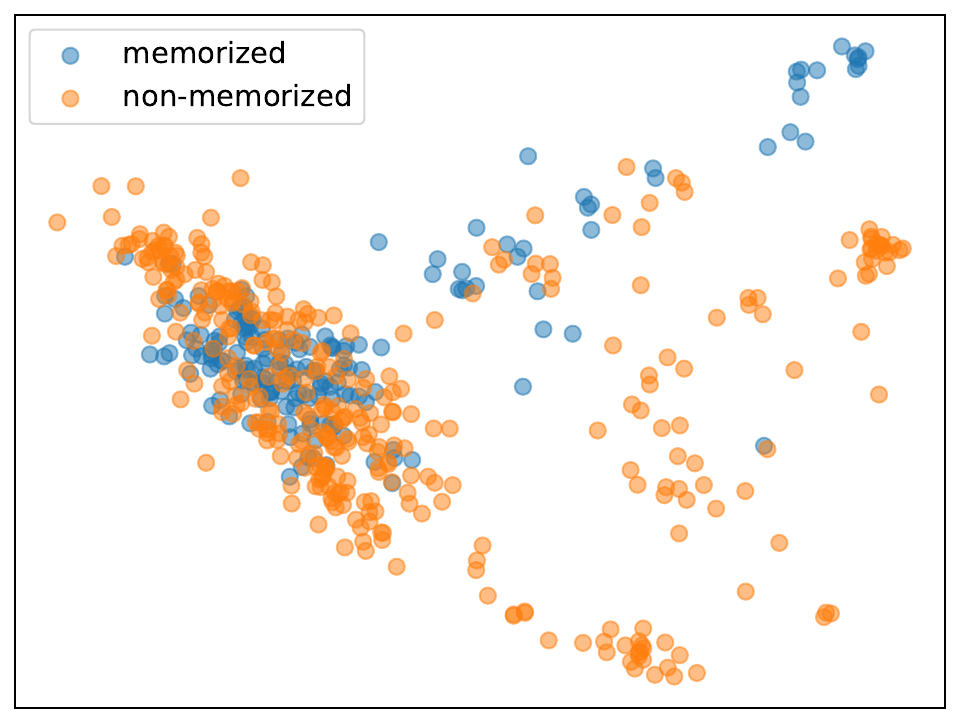}
  \includegraphics[width=0.16\linewidth]{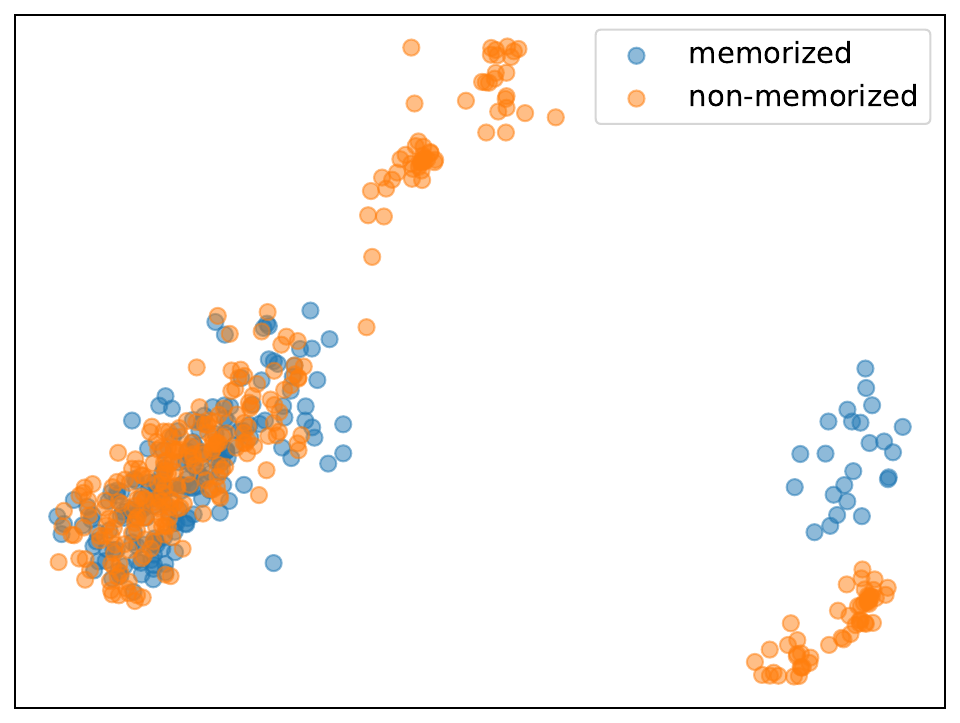}
  \includegraphics[width=0.16\linewidth]{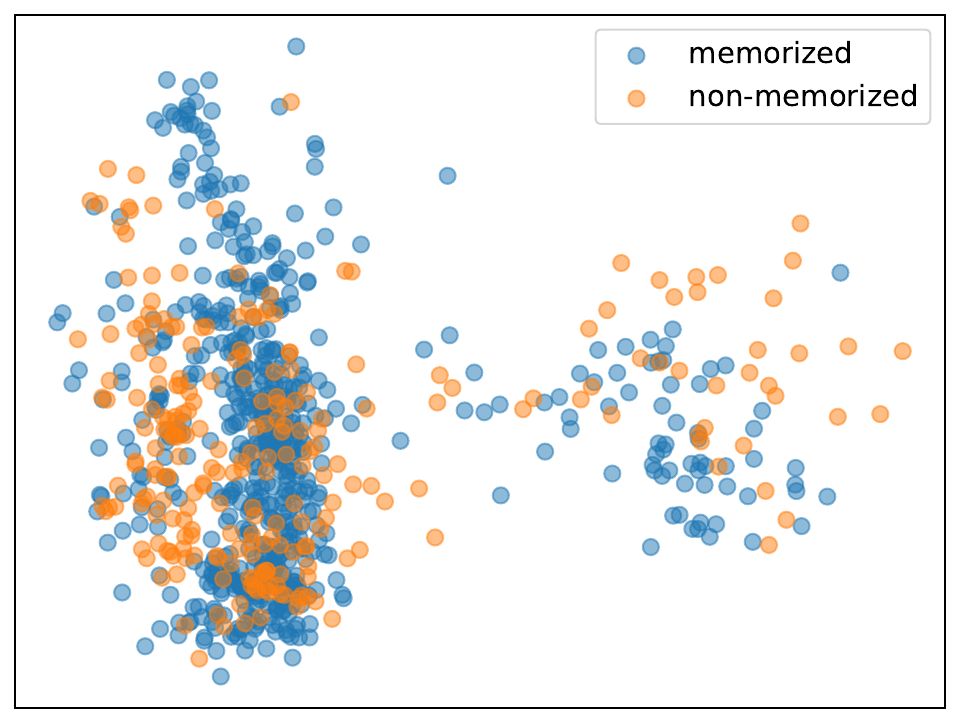}
  \hfill
  \includegraphics[width=0.16\linewidth]{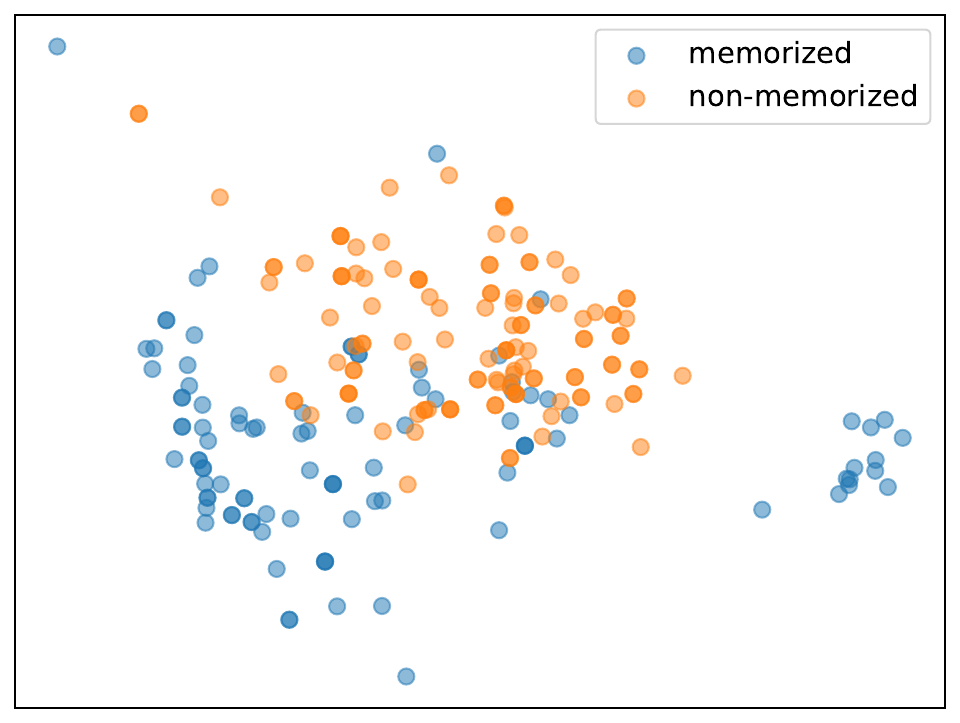}
  \includegraphics[width=0.16\linewidth]{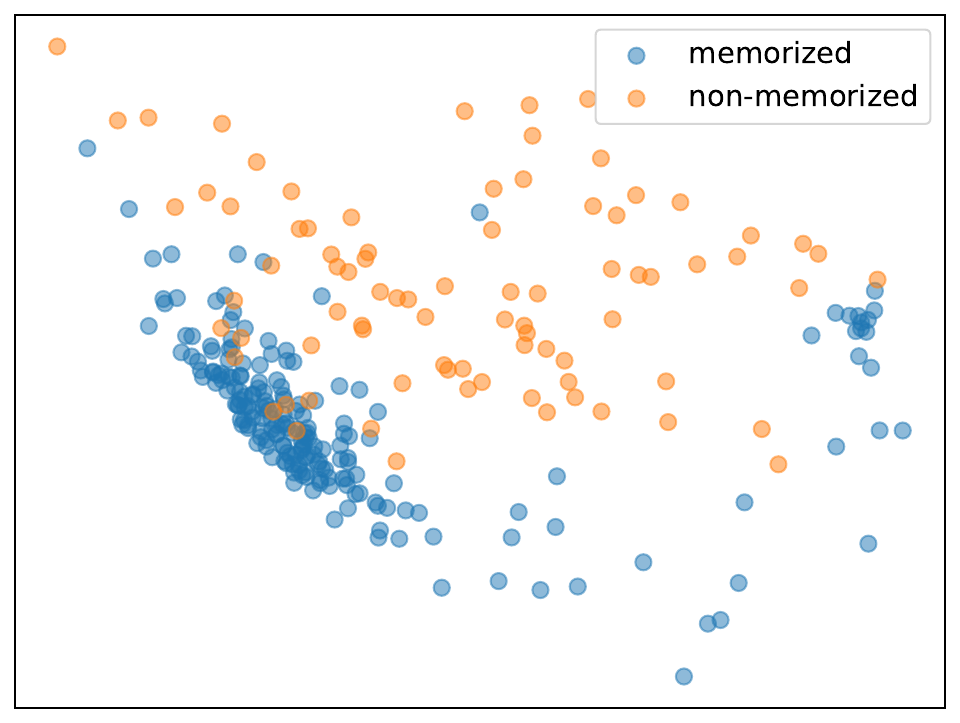}
  \includegraphics[width=0.16\linewidth]{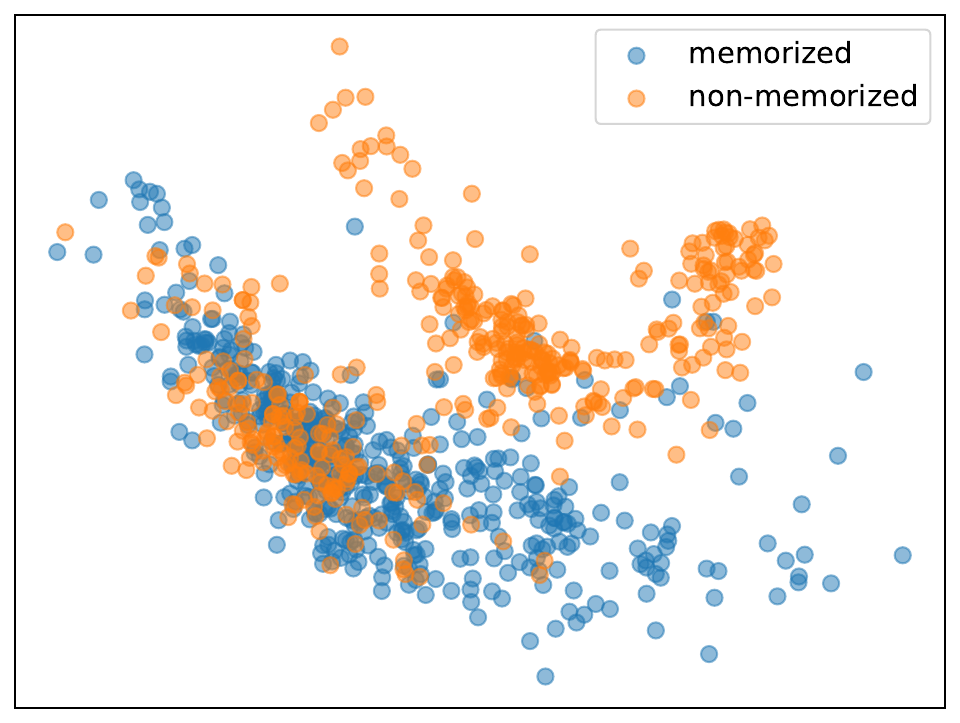}
  \includegraphics[width=0.16\linewidth]{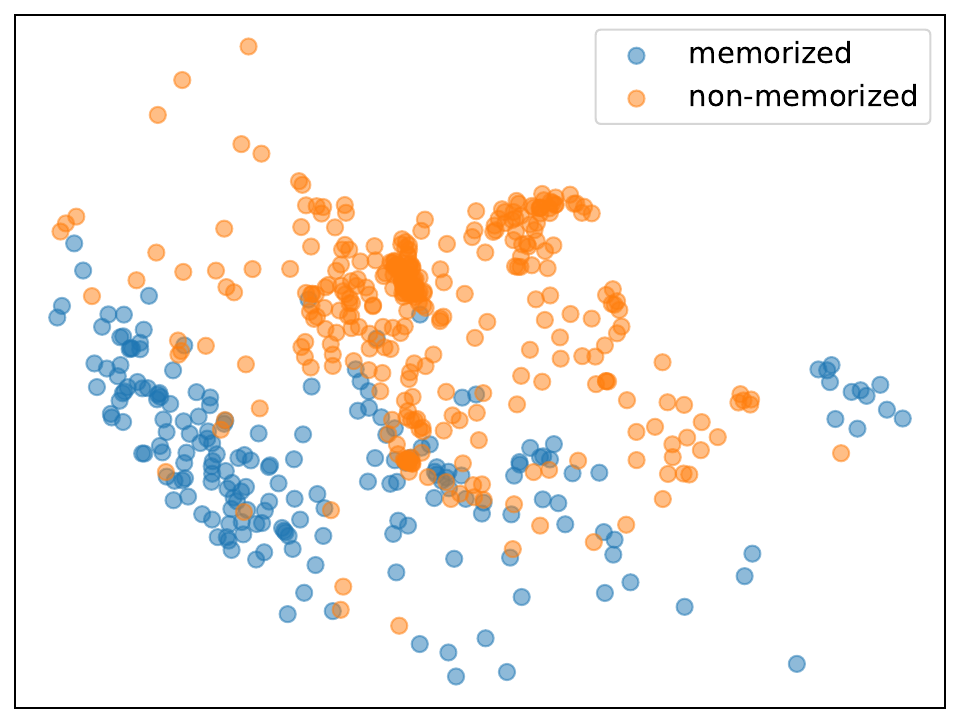}
  \includegraphics[width=0.16\linewidth]{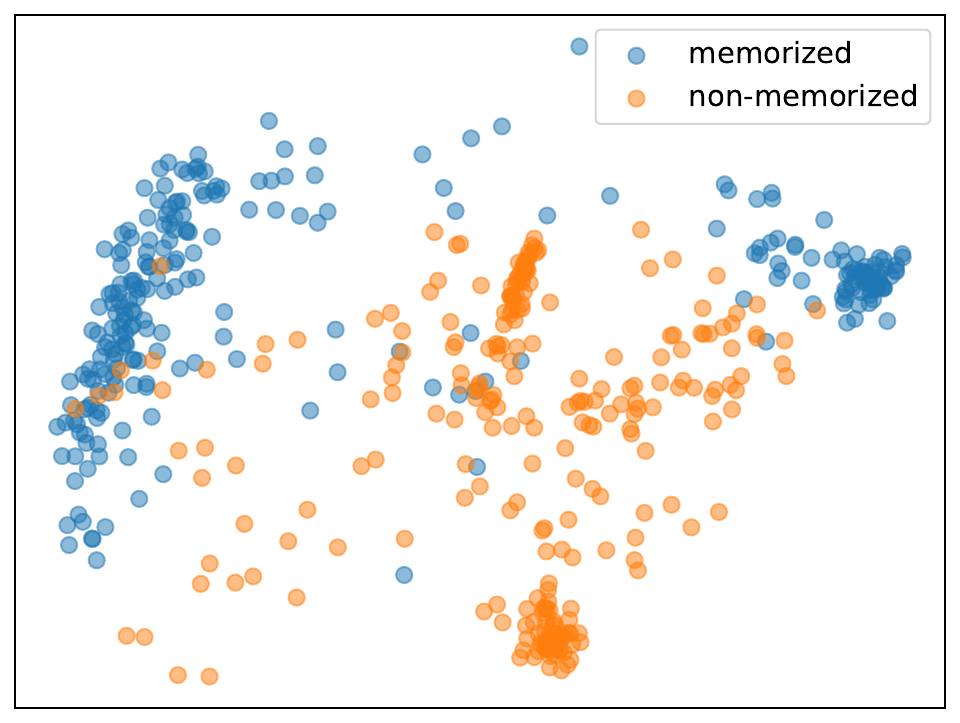}
  \includegraphics[width=0.16\linewidth]{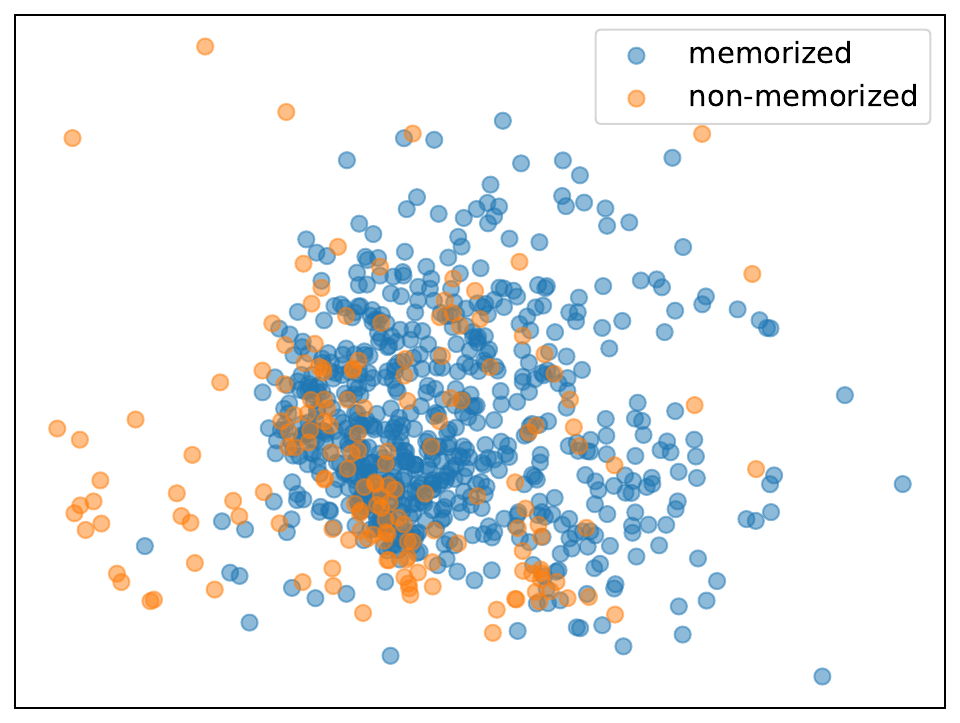}
  \hfill
  \includegraphics[width=0.16\linewidth]{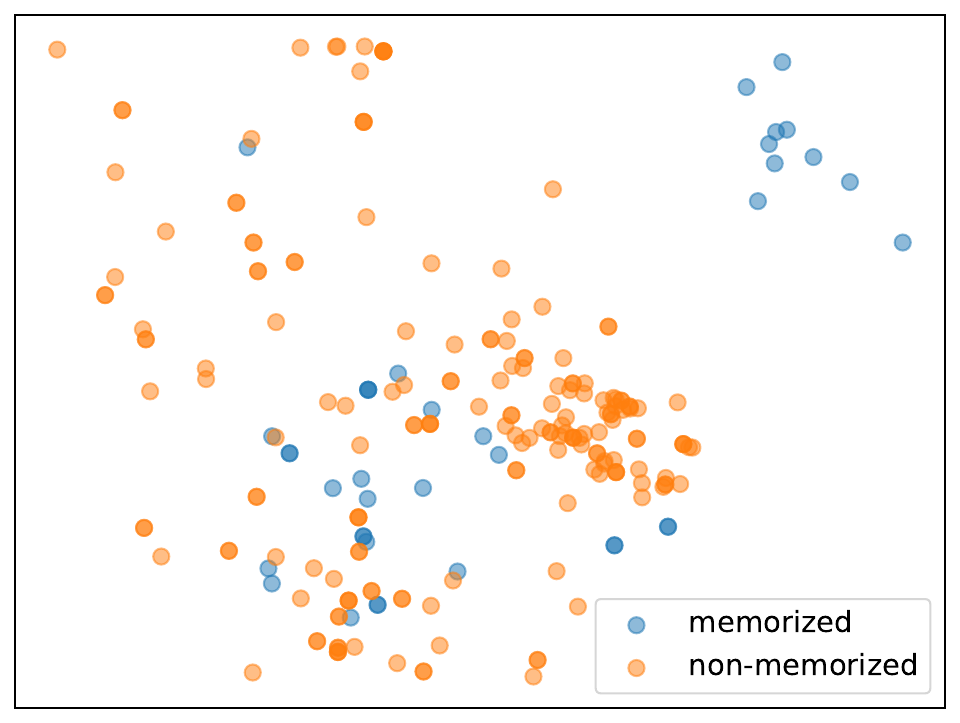}
  \includegraphics[width=0.16\linewidth]{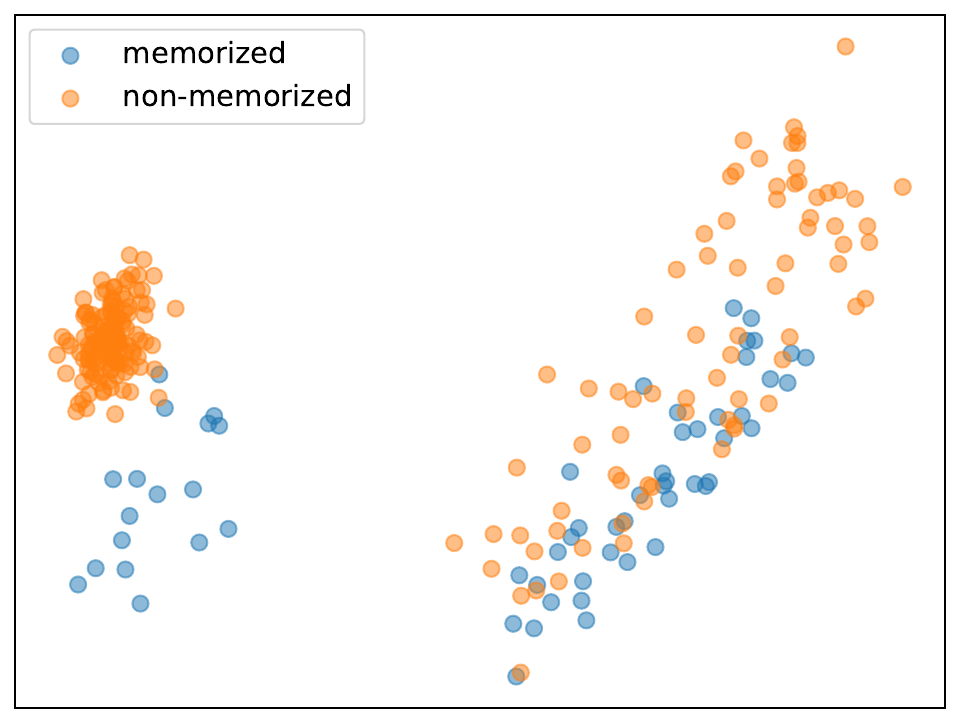}
  \includegraphics[width=0.16\linewidth]{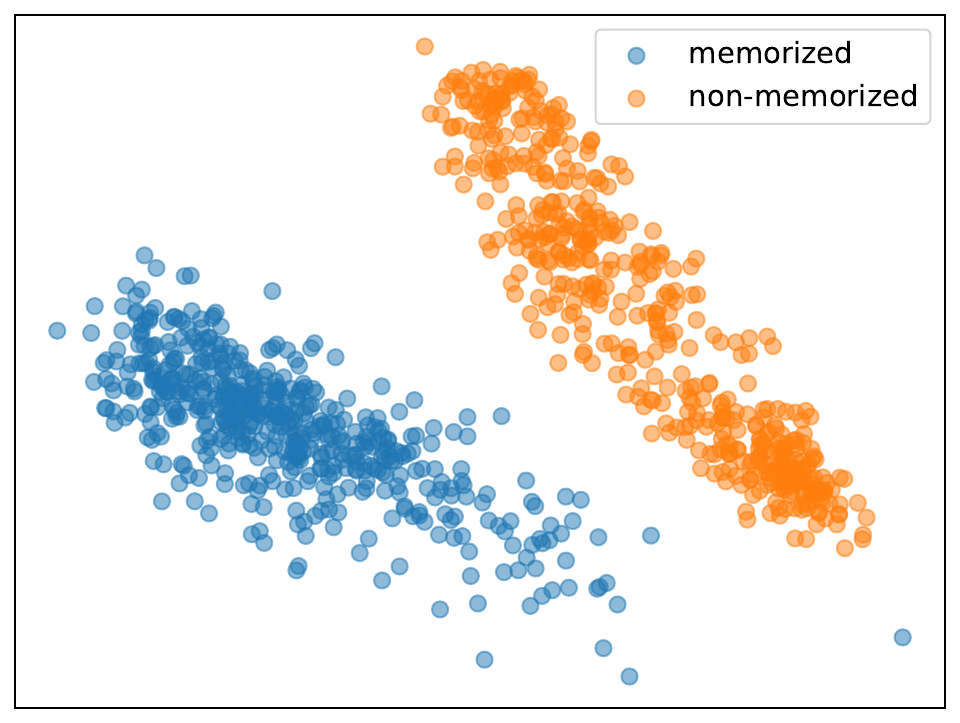}
  \includegraphics[width=0.16\linewidth]{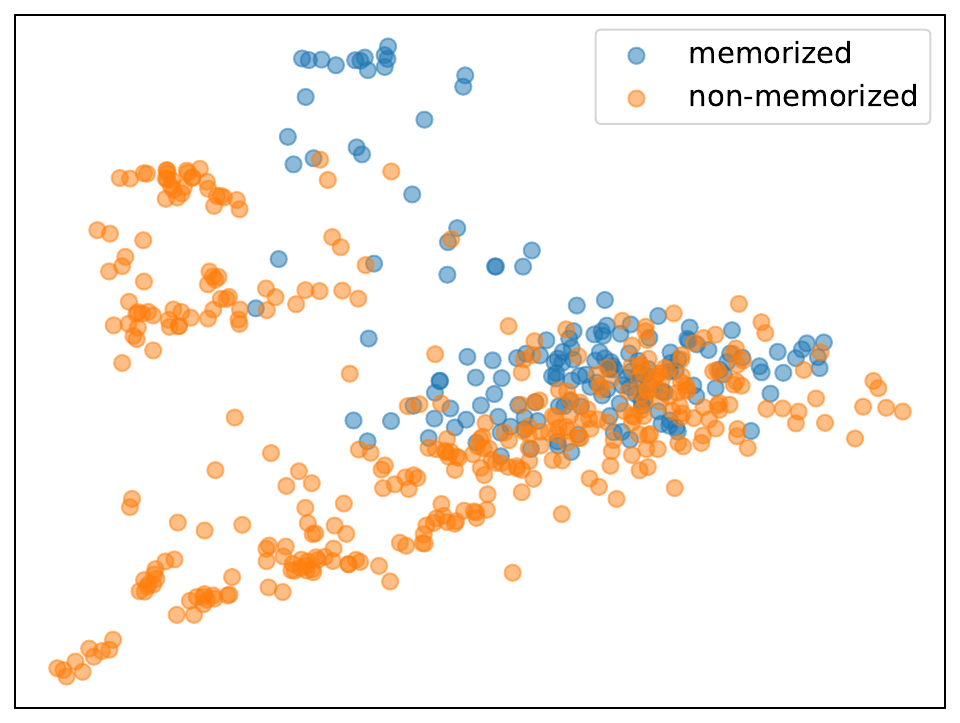}
  \includegraphics[width=0.16\linewidth]{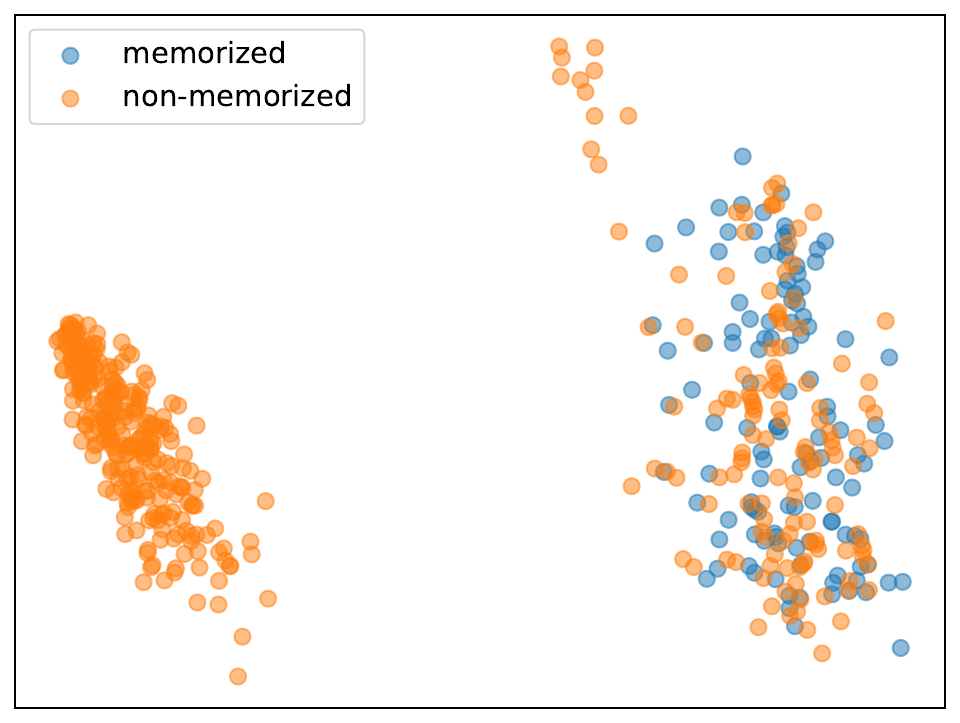}
  \includegraphics[width=0.16\linewidth]{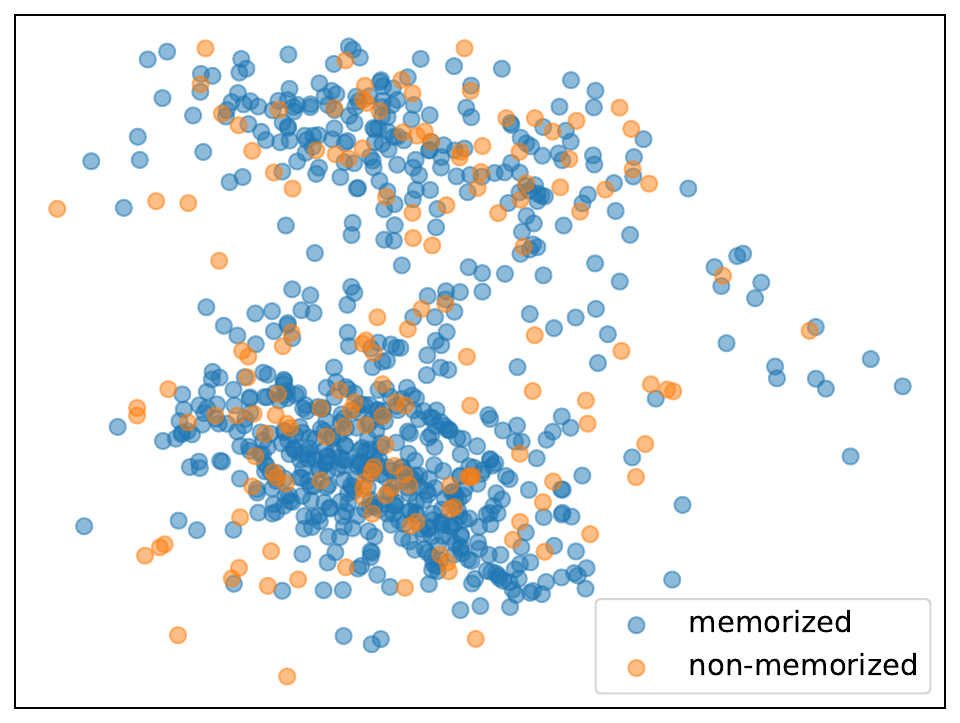}
  \hfill
  \includegraphics[width=0.16\linewidth]{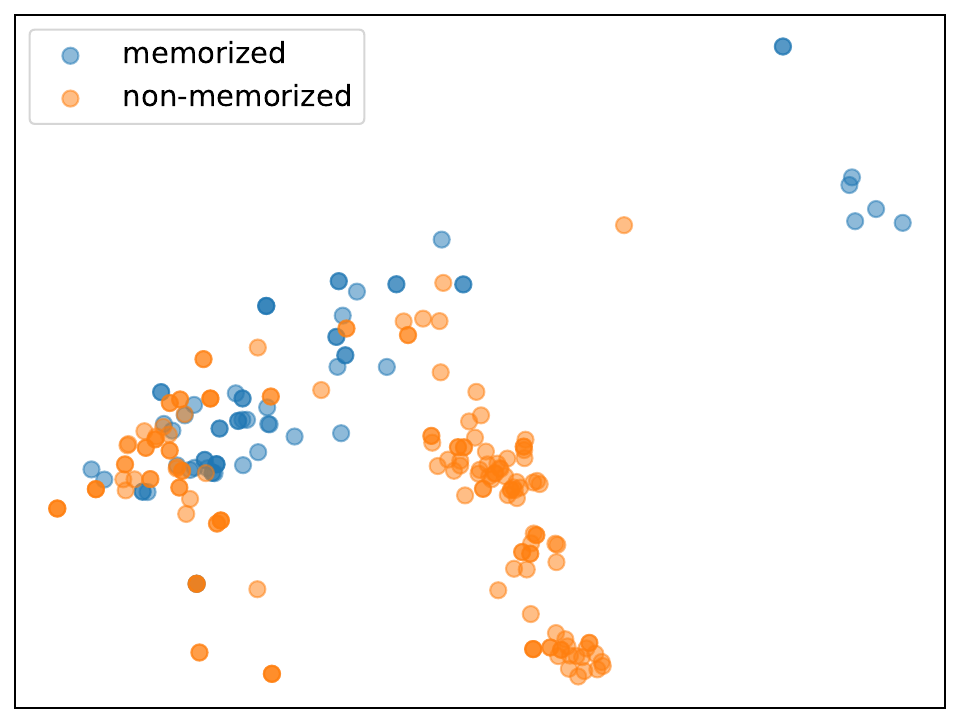}
  \includegraphics[width=0.16\linewidth]{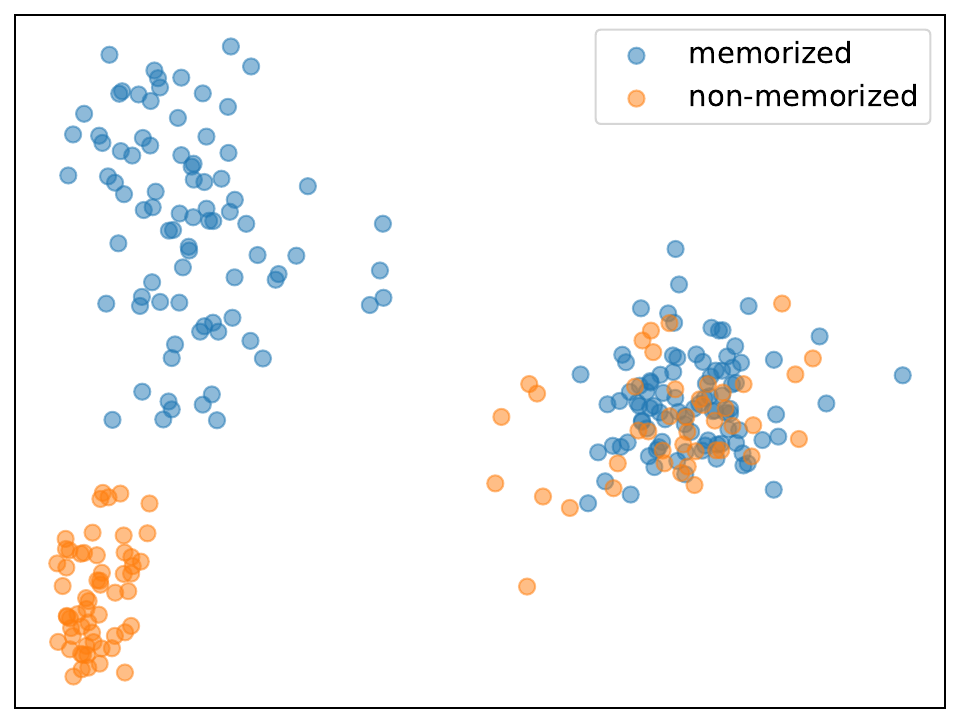}
  \includegraphics[width=0.16\linewidth]{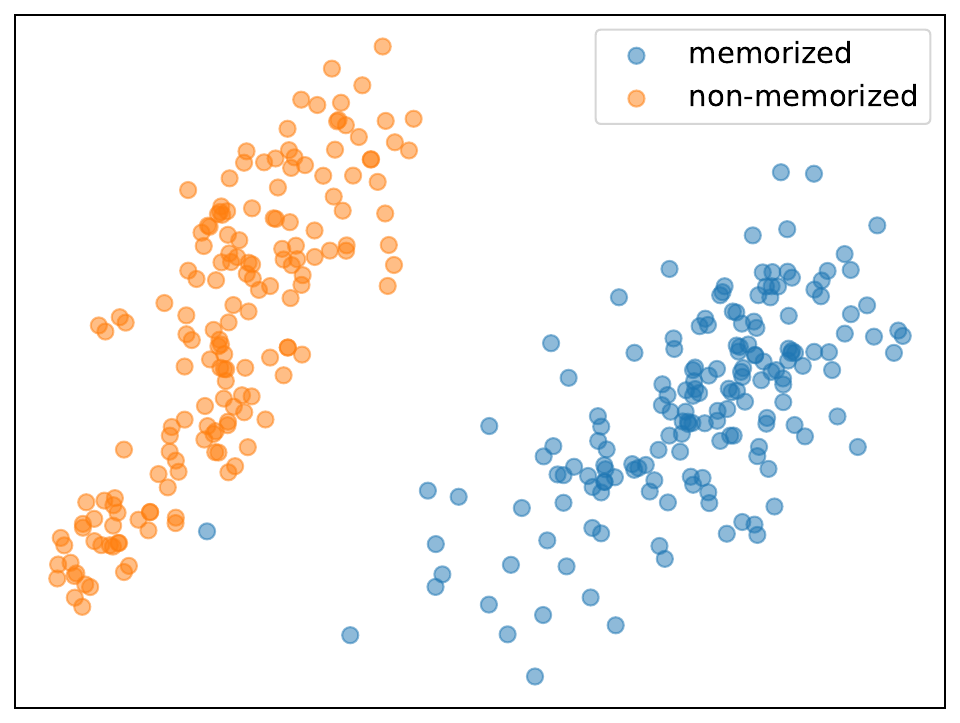}
  \includegraphics[width=0.16\linewidth]{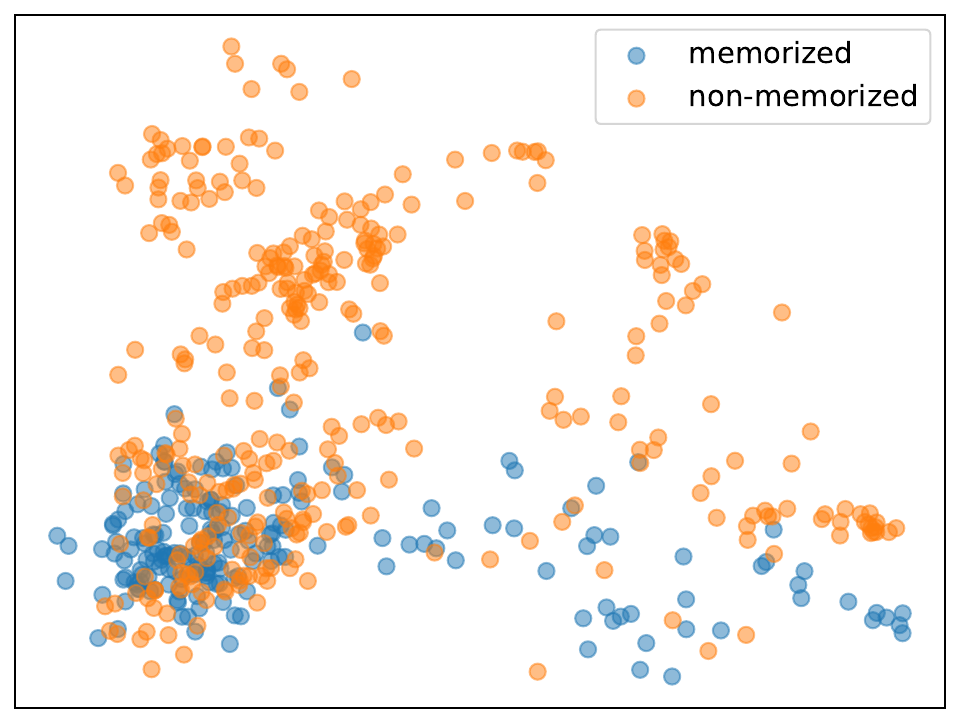}
  \includegraphics[width=0.16\linewidth]{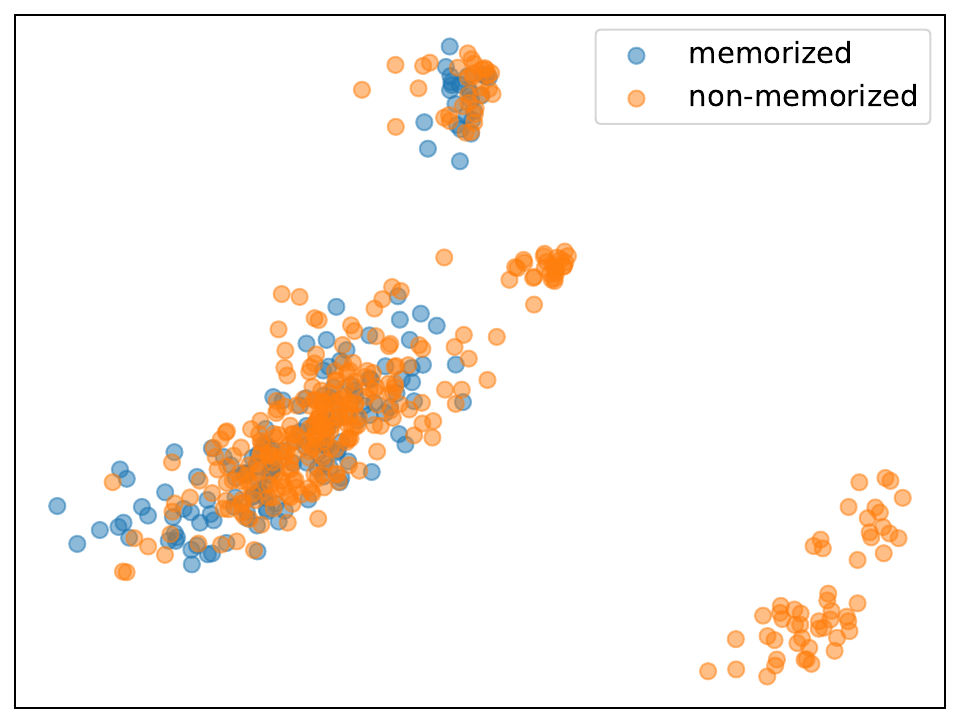}
  \includegraphics[width=0.16\linewidth]{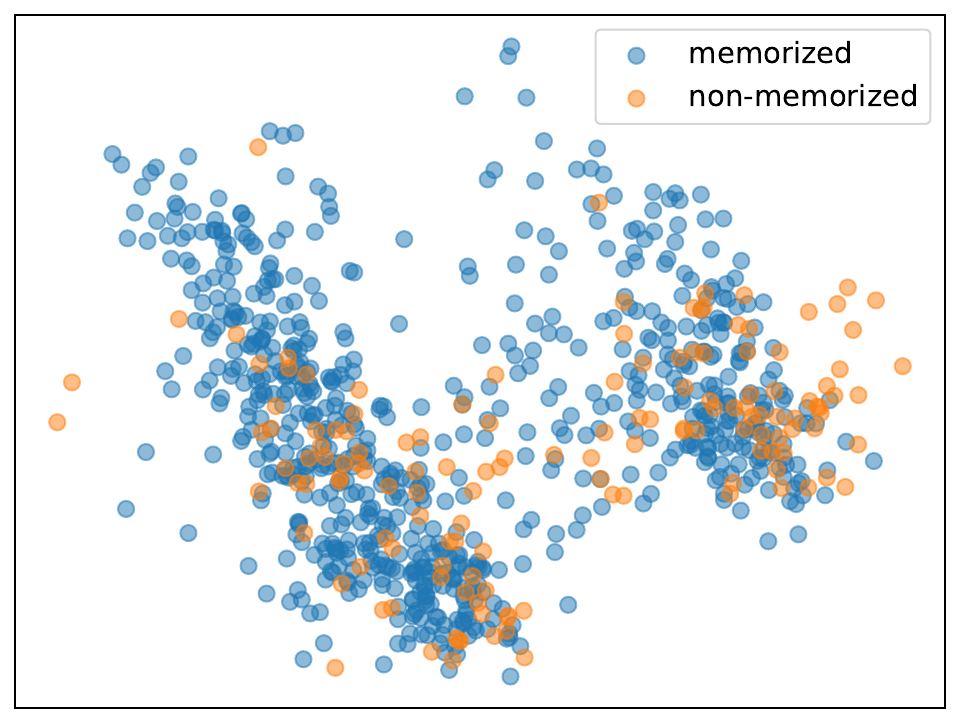}
  \hfill
  \includegraphics[width=0.16\linewidth]{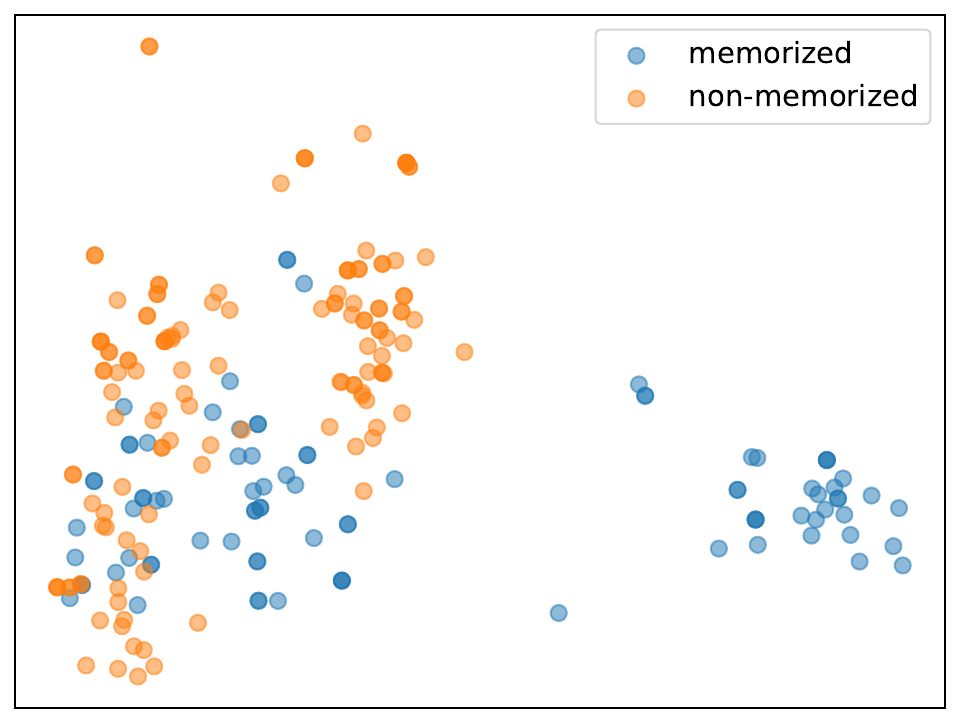}
  \includegraphics[width=0.16\linewidth]{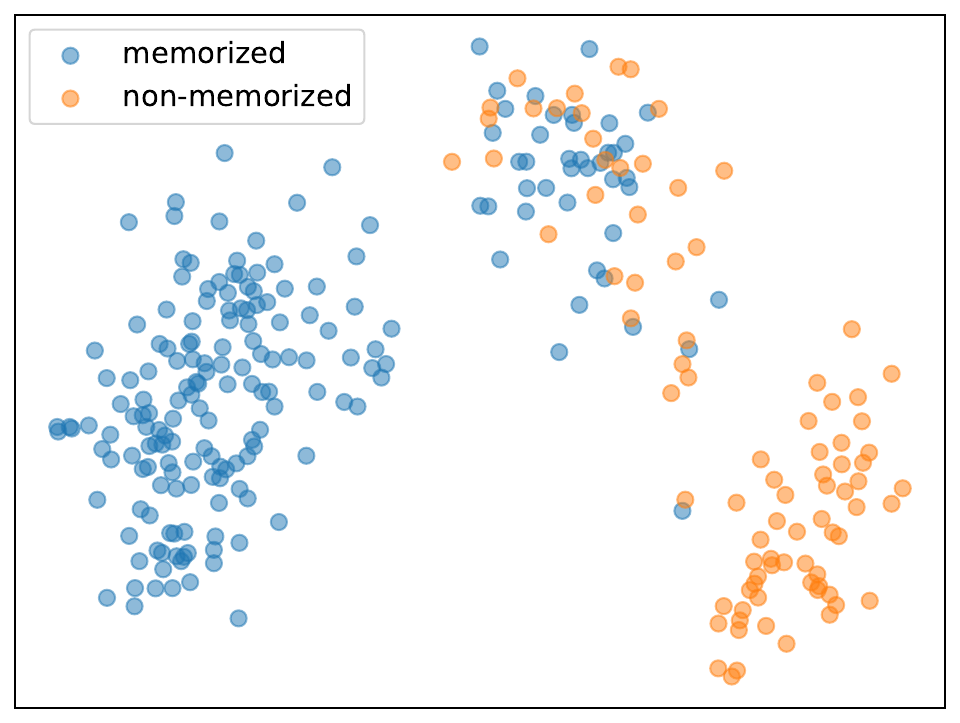}
  \includegraphics[width=0.16\linewidth]{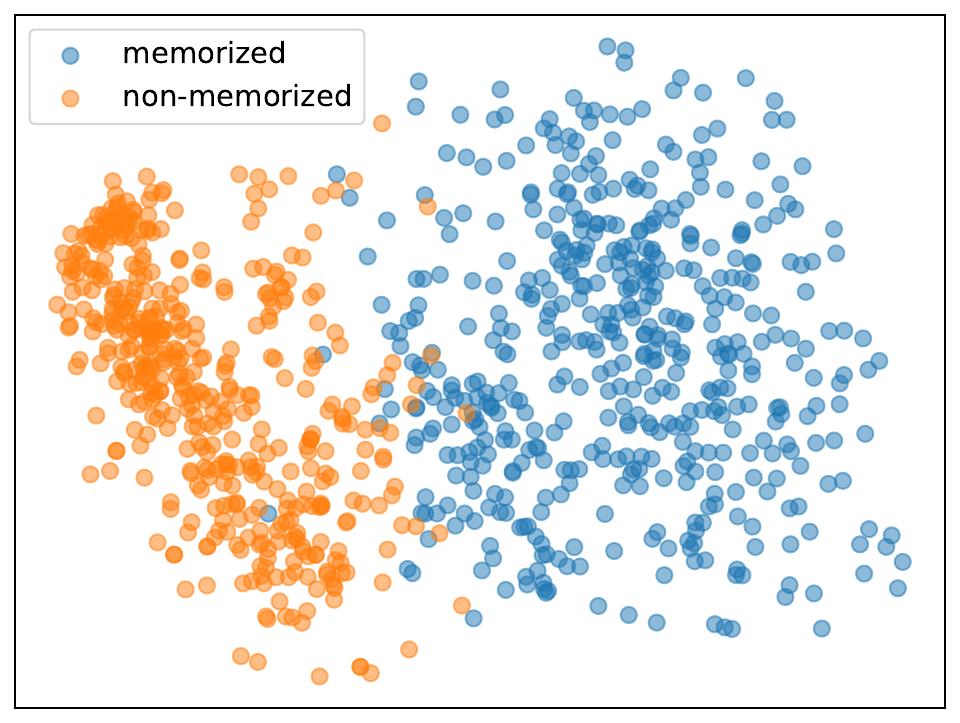}
  \includegraphics[width=0.16\linewidth]{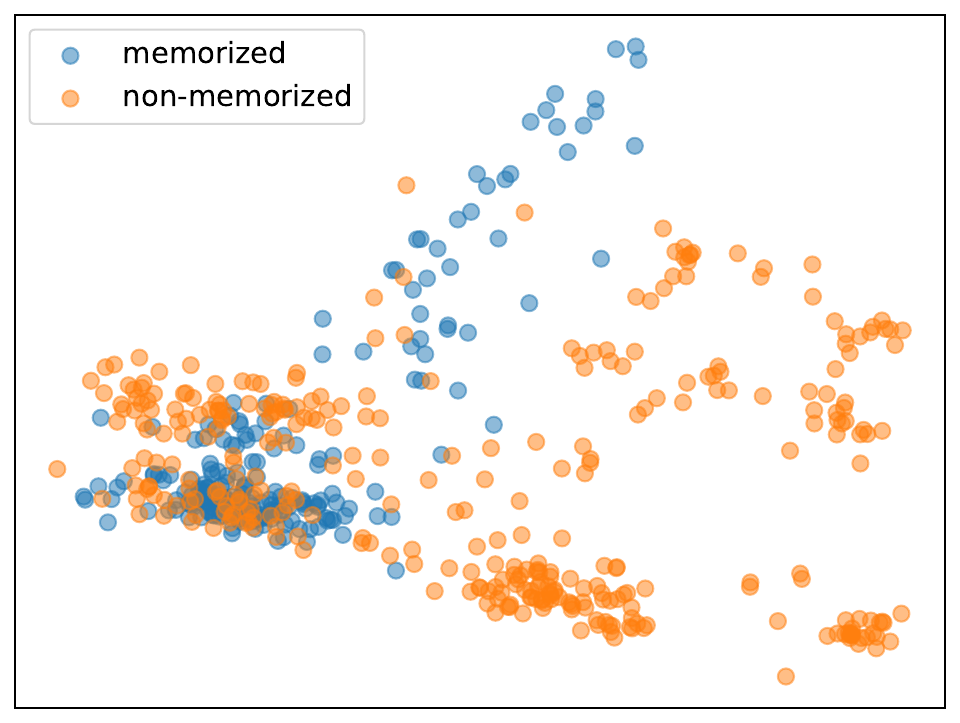}
  \includegraphics[width=0.16\linewidth]{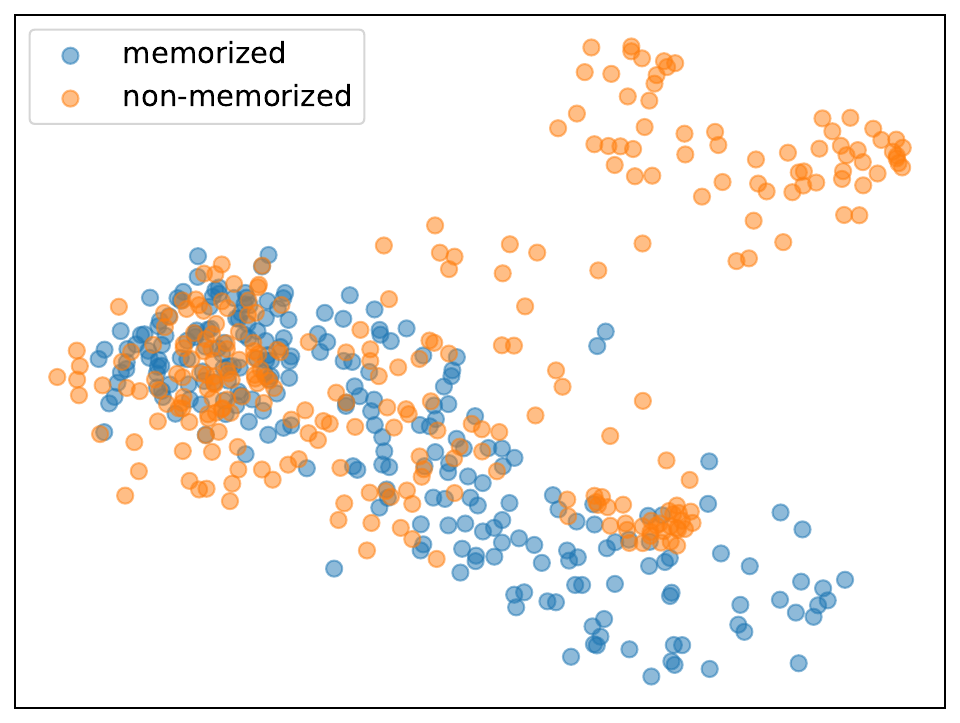}
  \includegraphics[width=0.16\linewidth]{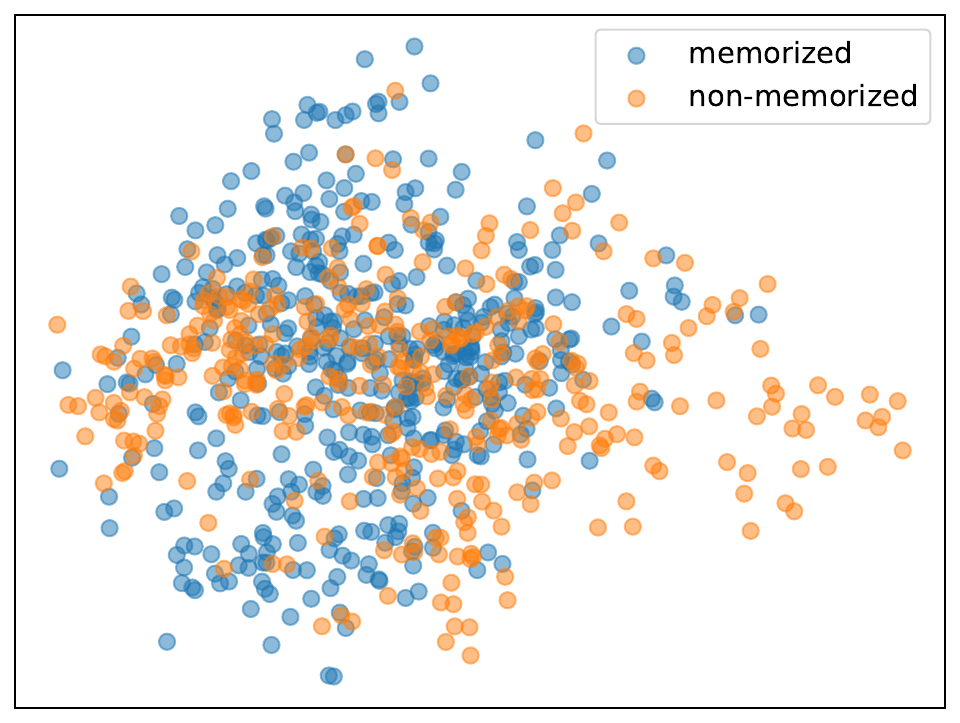}
  \caption{The visualization of representations grouped by \textit{memorized} and \textit{non-memorized} using PCA. From left to right, the dataset is \datasetterm, \datasetpn, \datasetcele, \datasetpopqa, \datasetlama~and \datasetidioms, respectively. For top to down, the model is \mistral~7B, \gemma~7B, \llama-2 7B, \llama-2 13B and \llama-3 8B, respectively.}
  \label{fig:pca_model_data}
\end{figure*}
\begin{figure}[t]
  \centering
    \includegraphics[width=0.75\linewidth]{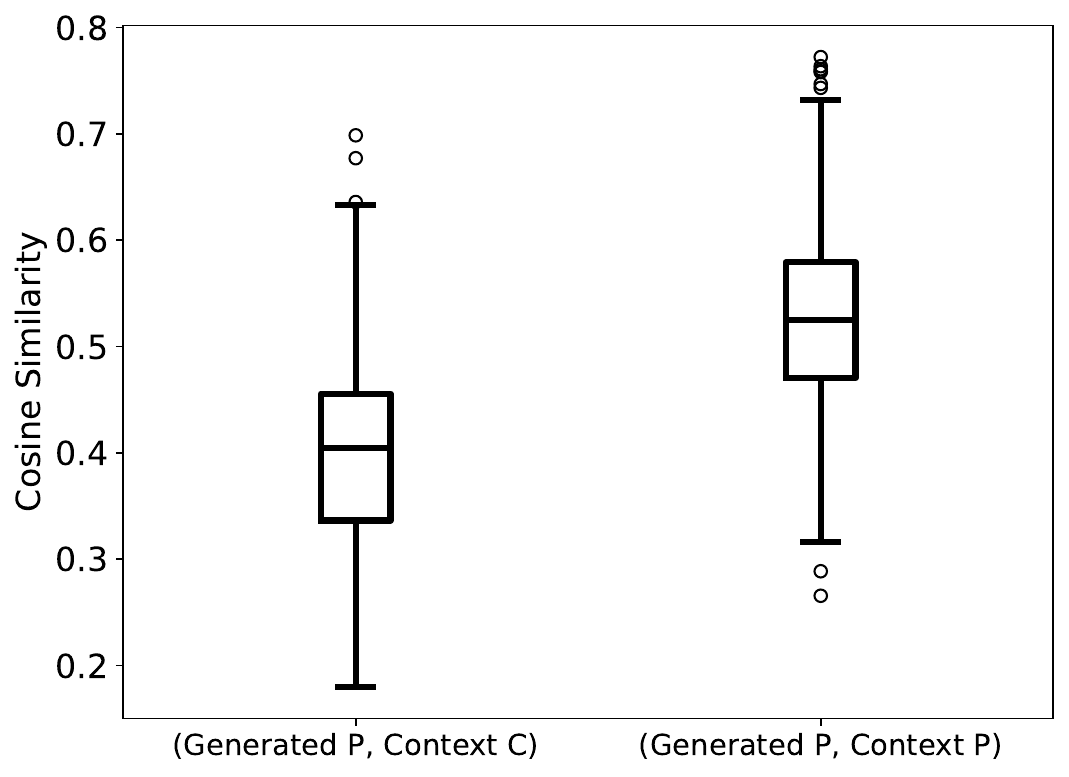}
  \caption{The demonstrations of the similarities between (Generated Parent, Context Child) and (Generated Parent, Context Parent). 
  }
  \label{fig:hidden_state_similarity}
\end{figure}
 
\subsection{Representation-oriented Analysis.}\label{sec:hidden_state}
The Section \ref{sec:text_oriented_analysis} and \ref{sec:probability_oriented_analysis} have demonstrated significant differences between memorized and non-memorized samples, with both generated text and logits derived from the representation. In this section, we explore the representations from both memorized and non-memorized samples.

\paragraph{Separated representations between \textit{memorized} and \textit{non-memorized}.}
Figure \ref{fig:pca_model_data} provides a visualization using PCA dimensionality reduction of vectors for both memorized (blue points) and non-memorized (yellow points) groups. Notably, while there is some overlap between the two groups, they are generally distinctly separated in most cases, suggesting that their representations in the vector space are relatively unique. This distinction is consistently observed across different models. However, significant variations are evident between different datasets. For example, data points in \datasetcele, \datasetpopqa, and \datasetlama~tend to cluster more tightly, whereas those in \datasetpn~and \datasetterm show higher levels of dispersion. In the case of \datasetidioms, the distinction between memorized and non-memorized groups is less apparent. 
Additionally, the probability values across this dataset are generally low, and the difference in probability values between the two groups is minimal (as shown in Table \ref{tab:probability_analysis}). It is hypothesized that these two phenomena is caused by the same factors.

\paragraph{Same concepts are more similar.}
To determine whether the representation of the same concept is more similar across different contexts than other concepts within the same context, we focus on the dataset \datasetcele, which encapsulates an inverse relationship between parents and children. We analyze memorization through the similarities of representations between these related names. 
Specifically, for the query ``Who is \textit{Elon Musk}'s mother?'' with the response ''Maye Musk'',    ``Elon Musk'' is referred to as the ``Context Child'' since his name appears in the query context, and ``Maye Musk'' is designated as the ``Generated Parent'' as it is produced by the model. Conversely, for the query ``Name a child of Maye Musk'',  ``Maye Musk'' is labeled as the ``Context Parent''.
 To quantify these similarities, we use cosine similarity, the most commonly employed metric for assessing vector similarity. The results, depicted in Figure \ref{fig:hidden_state_similarity} and presented solely for the \llama-2 13B model due to clarity and space constraints, demonstrate that representations of the same concepts are more similar to each other than to different concepts in similar contexts. 
 This to some extent indicates that memorization, rather than context, influences the representation of these names more.

Additionally, as noted in \citep{berglund2023reversal}, models frequently correctly answer ``Who is Elon Musk's mother?'', but struggle with ``Name a child of Maye Musk''. These findings indicate distinct memorization capabilities for the same names, even though they share equal word frequency. This suggests that memorization is not merely determined by the frequency of a concept's occurrence.

\section{Conclusion}
We developed \modelname~to explore the memorization processes in LLMs, focusing on a comparative analysis between memorized and non-memorized groups rather than directly accessing the training data. To ensure that the memorized group aligns more closely with the conventional understanding of memorization, we carefully selected datasets that exhibit the ``block in block out'' characteristics. In conjunction with factual question answering datasets, we investigated the model behavior patterns of these groups across seven datasets, examining dimensions such as text, logits, and representation. This study has uncovered several intriguing findings, such as models produce higher confidence for memorized samples.

\section*{Limitations}
In this work, we conduct comparative analysis from the insights of text, probability and representation. Although our study yields several intriguing empirical findings, some limitations still exist.
Firstly, the empirical findings have not yet been fully validated in larger models, 
such as \llama-3 70B. Secondly, other factors, including the used prompts and the explored insights, can influence the memorization characteristics of LLMs.

\bibliography{refs}
\appendix
\section{Appendix}
\label{sec:appendix}

\subsection{Examples}\label{sec:examples}
\begin{table*}[ht]
\centering
{\renewcommand{\arraystretch}{1.2}
\resizebox{0.8\textwidth}{!}{
\begin{tabular}{l|l}
\toprule 
\textbf{Dataset} & \textbf{Examples} \\
\toprule 
\multirow{3}{*}{\datasetidioms} & no pain no gain \\
 & see eye to eye \\
 & in the wink of an eye \\
\hline
\multirow{3}{*}{\datasettp} & \zhsmall{红颜未老恩先断}\\
 & \zhsmall{十年一觉扬州梦} \\
 &\zhsmall{千山鸟飞绝}\\
 
\hline
\multirow{3}{*}{\datasetpn} & Royal College of Art \\
 & Indian Institute of Technology \\
 & Potsdam Institute for Climate Impact Research \\
\hline
\multirow{3}{*}{\datasetterm} & Acquired Immunodeficiency Syndrome \\
 & Mixed Connective Tissue Disease \\
 & Toxic Epidermal Necrolysis \\
\hline
\multirow{3}{*}{\datasetpopqa} & Q: What is Javad Maroufi's occupation? A: composer \\
 & Q: What sport does Richard Varga play? A: triathlon \\
 & Q: What is the capital of Australia? A: Canberra \\
\hline
\multirow{3}{*}{\datasetlama} & Q: Kaare Fostervoll died in July 1981 in [MASK]. A: Oslo \\
 & Q: In 1711, being very ill, Aveneau retired to [MASK]. A: Quebec \\
 & Q: McLean died on 30 March 2011 at his home in [MASK]. A: Wellington \\
\hline
\multirow{3}{*}{\parbox{4cm}{\datasetcele\\ (parents prediction)}} & Who is Tom Cruise's mother? Mary Lee Pfeiffer\\
&Who is Jennifer Lawrence's mother? Karen Lawrence\\
&Who is Hailee Steinfeld's father? Peter Steinfeld\\
\hline
\multirow{3}{*}{\parbox{4cm}{\datasetcele\\ (child prediction)}} &Name a child of Peter Steinfeld? Hailee Steinfeld\\
&Name a child of Joy Vogelsang? Nicolas Cage\\
&Name a child of Sarah Johnson? Aaron Taylor-Johnson\\
\bottomrule 
\end{tabular}
}
}
\caption{Examples of each dataset.}
\label{example_datasets}
\end{table*}

\subsection{Exemplars}\label{sec:exemplars}

\begin{table*}[ht]
\centering
{\renewcommand{\arraystretch}{1.2}
\resizebox{0.8\textwidth}{!}{
\begin{tabular}{p{0.8\linewidth}}
\toprule
\textbf{[\datasetidioms, zero-shot]} \\
\toprule
PROMPT: it doesn't hurt to \_\_  \\
ANSWER: {ask} \\
\toprule
\textbf{[\datasettp, zero-shot]} \\
\toprule
PROMPT: \zhsmall{千山鸟飞\_\_} \\
ANSWER: \zhsmall{绝}\\ 
\bottomrule 
\end{tabular}
}
}
\caption{Exemplars of \datasetidioms~and \datasettp.}
\label{prmopt1}
\end{table*}
\begin{table*}[ht]
\centering
{\renewcommand{\arraystretch}{1.2}
\resizebox{0.8\textwidth}{!}{
\begin{tabular}{p{0.8\linewidth}}
\toprule
\textbf{[\datasetpn, 8-shot]} \\
\bottomrule
Saint Catherine's \_\_ Sinai\\
Monastery\\
Alfred P. \_\_ Foundation\\
Sloan\\
Center for \_\_ Dynamics\\
Nonlinear\\
The Bank of New \_\_ Mellon\\
York\\
Greenland Institute of \_\_ Resources\\
Natural\\
Smithsonian American \_\_ Museum\\
Art\\
Financial Times \_\_ Exchange\\
Stock\\
Yale School \_\_ Drama\\
of\\
\toprule
\textbf{[\datasetterm, 8-shot]} \\
\bottomrule
Chronic Obstructive \_\_ Disease\\
Pulmonary\\
Acute \_\_ Leukemia\\
Lymphoblastic\\
Acute \_\_ Leukemia\\
Myeloid\\
Chronic \_\_ Syndrome\\
Fatigue\\
Rhizomelic \_\_ Punctata\\
Chondrodysplasia\\
Sanfilippo Syndrome \_\_ A\\
Type\\
Severe \_\_ Immunodeficiency\\
Combined\\
Spinal \_\_ Atrophy\\
Muscular\\
\bottomrule 
\end{tabular}
}
}
\caption{Exemplars of \datasetpn~and \datasetterm.}
\label{prmopt2}
\end{table*}

\begin{table*}[ht]
\centering
{\renewcommand{\arraystretch}{1.2}
\resizebox{0.9\linewidth}{!}{
\begin{tabular}{p{0.9\linewidth}}
\toprule
\textbf{[\datasetpopqa, 5-shot]} \\
\toprule
Q: What is Bruce McDaniel's occupation?  
A: composer  \\ 
Q: What is William Lescaze's occupation?  
A: architect  \\
Q: What is Tsutomu Seki's occupation?  
A: astronomer  \\
Q: What is Dominick Bellizzi's occupation?  
A: jockey  \\
Q: What is Michael Arad's occupation? 
A: architect \\
\bottomrule
\textbf{[\datasetlama, 4-shot]} \\
\toprule
Q: Paul Mounsey (born 15 April 1959) is a composer arranger and producer from [MASK]. \\
A: Scotland \\ 
Q: Pierre Dupont (April 23 1821 – July 25 1870) French song-writer the son of a blacksmith was born in [MASK]. \\
A: Lyon \\
Q: Susette La Flesche (later Susette LaFlesche Tibbles) also called Inshata Theumba (Bright Eyes) (1854 – 1903) was a well-known Native American writer lecturer interpreter and artist of the Omaha tribe in [MASK]. \\
A: Nebraska \\
Q: The revolt of Husayn ibn Ali ibn Hasan broke out when Husayn declared himself caliph in [MASK].\\
A: Medina \\
\bottomrule
\textbf{[\datasetcele, 6-shot]} \\
\toprule
Q: Name a child of Barack Obama. 
A: Malia Obama \\
Q: Who is Elon Musk's mother? 
A: Maye Musk \\
Q: Name a child of Donald Trump. 
A: Ivanka Trump \\
Q: Who is Chris Hemsworth's father? 
A: Craig Hemsworth \\
Q: Name a child of Karen Lawrence. 
A: Jennifer Lawrence \\
Q: Who is Aaron Taylor-Johnson's mother? 
A: Sarah Johnson \\
\bottomrule 
\end{tabular}
}
}
\caption{Exemplars of \datasetpopqa, \datasetlama~and \datasetcele.}
\label{prmopt3}
\end{table*}
\end{document}